\documentclass{article} 
\usepackage{iclr2023_conference,times}

\usepackage[utf8]{inputenc} 
\usepackage[T1]{fontenc}    
\usepackage[pagebackref=true,breaklinks=true,colorlinks,bookmarks=false]{hyperref}       
\usepackage{url}            
\usepackage{booktabs}       
\usepackage{amsfonts}       
\usepackage{nicefrac}       
\usepackage{microtype}      
\usepackage{graphicx}
\usepackage{caption}
\usepackage{anyfontsize}

\usepackage{floatrow}
\usepackage{bm}
\usepackage[normalem]{ulem}
\usepackage{natbib}
\usepackage{xcolor, colortbl}
\usepackage{xspace}
\usepackage{lipsum}
\usepackage{siunitx}
\usepackage{subfigure}
\usepackage{multirow}
\usepackage[font=small,labelfont=bf]{caption}
\newfloatcommand{capbtabbox}{table}[][\FBwidth]

\usepackage{amsmath,amsfonts,bm}

\def\eqref#1{equation~\ref{#1}}

\def\1{\bm{1}}

\def\vtheta{{\bm{\theta}}}
\def\vphi{{\bm{\phi}}}
\def\vpsi{{\bm{\psi}}}
\def\vdelta{{\bm{\delta}}}

\def\vx{{\bm{x}}}

\DeclareMathAlphabet{\mathsfit}{\encodingdefault}{\sfdefault}{m}{sl}
\SetMathAlphabet{\mathsfit}{bold}{\encodingdefault}{\sfdefault}{bx}{n}

\def\sS{{\mathbb{S}}}

\DeclareMathOperator*{\maximize}{max}

\definecolor{header}{gray}{1}
\definecolor{subheader2}{rgb}{1, 0.9, 0.8}
\definecolor{subheader}{rgb}{0.87, 1, 0.82}
\newcommand{\Tstrut}{\rule{0pt}{2.6ex}}
\newcommand{\Bstrut}{\rule[-0.9ex]{0pt}{0pt}}
\newcommand{\TBstrut}{\Tstrut\Bstrut}

\usepackage{algorithm}
\usepackage{algpseudocode}
\usepackage{algorithmicx}
\algrenewcommand\algorithmicrequire{\textbf{Input:}}
\algrenewcommand\algorithmicensure{\textbf{Output:}}

\usepackage{glossaries}
\usepackage[inline]{enumitem}
\glsdisablehyper

\definecolor{TartOrange}{HTML}{ff2e35}
\definecolor{Orange}{HTML}{ff7825}
\definecolor{Mango}{HTML}{ffc013}
\definecolor{AppleGreen}{HTML}{7cb81b}
\definecolor{Blue}{HTML}{1173b0}
\definecolor{BdazzledBlue}{HTML}{2e58a5}
\definecolor{Purple}{HTML}{5b3590}
\definecolor{Sunglow}{HTML}{FFCA3A}
\definecolor{TableRow}{gray}{0.9}

\newcommand{\imagenet}{\textsc{ImageNet}\xspace}
\newcommand{\imageneta}{\textsc{ImageNet-A}\xspace}
\newcommand{\imagenetc}{\textsc{ImageNet-C}\xspace}
\newcommand{\imagenetr}{\textsc{ImageNet-R}\xspace}
\newcommand{\imagenetsketch}{\textsc{ImageNet-Sketch}\xspace}
\newcommand{\mnist}{\textsc{Mnist}\xspace}
\newcommand{\cifar}{\textsc{Cifar-10}\xspace}
\newcommand{\cifarh}{\textsc{Cifar-100}\xspace}
\newcommand{\resnet}{\textsc{ResNet}\xspace}
\newcommand{\vit}{\textsc{ViT}\xspace}
\newcommand{\vitb}{\textsc{ViT-B16}\xspace}
\newcommand{\vits}{\textsc{ViT-S}\xspace}

\newcommand{\linf}{\ensuremath{\ell_\infty}\xspace}
\newcommand{\ltwo}{\ensuremath{\ell_2}\xspace}

\newcommand{\pgd}[1]{\textsc{PGD}\textsuperscript{$#1$}\xspace}
\newcommand{\autoattack}{\textsc{AutoAttack}\xspace}
\newcommand{\multitargeted}{\textsc{MultiTargeted}\xspace}

\newacronym{resnet}{\resnet}{Residual Network}

\newcommand{\squishlist}{
   \begin{list}{$\bullet$}
    {\setlength{\itemsep}{0pt} \setlength{\parsep}{3pt}
     \setlength{\topsep}{3pt} \setlength{\partopsep}{0pt}
     \setlength{\leftmargin}{1em} \setlength{\labelwidth}{1em}
     \setlength{\labelsep}{0.5em}}}
\newcommand{\squishend}{
    \end{list}}

\usepackage[framemethod=tikz]{mdframed}
\mdfdefinestyle{mystyle}{%
  linecolor=black!80,
  fontcolor=black!80,
  leftline=true,
  innerleftmargin=10,
  innerrightmargin=0,
  outerlinewidth=3pt,
  topline=false,
  rightline=false,
  bottomline=false,
  skipabove=\topsep,
  skipbelow=\topsep
}

\title{Revisiting adapters with adversarial training}

\author{Sylvestre-Alvise Rebuffi, Francesco Croce\thanks{Work done during an internship at DeepMind} ~\& Sven Gowal \\
DeepMind, London\\
\texttt{\{sylvestre,sgowal\}@deepmind.com}
\vspace{-0.1cm}
}

\iclrfinalcopy 
\begin{document}

\maketitle

\begin{abstract}
While adversarial training is generally used as a defense mechanism, recent works show that it can also act as a regularizer.
By co-training a neural network on clean and adversarial inputs, it is possible to improve classification accuracy on the clean, non-adversarial inputs.
We demonstrate that, contrary to previous findings, it is not necessary to separate batch statistics when co-training on clean and adversarial inputs, and that it is sufficient to use adapters with few domain-specific parameters for each type of input.
We establish that using the classification token of a Vision Transformer (\vit) as an adapter is enough to match the classification performance of dual normalization layers, while using significantly less additional parameters.
First, we improve upon the top-1 accuracy of a non-adversarially trained \vitb model by +1.12\% on \imagenet (reaching 83.76\% top-1 accuracy).
Second, and more importantly, we show that training with adapters enables \emph{model soups} through linear combinations of the \emph{clean} and \emph{adversarial} tokens.
These \emph{model soups}, which we call \emph{adversarial model soups}, allow us to trade-off between clean and robust accuracy without sacrificing efficiency.
Finally, we show that we can easily adapt the resulting models in the face of distribution shifts.
Our \vitb obtains top-1 accuracies on \imagenet variants that are on average +4.00\% better than those obtained with Masked Autoencoders.

\end{abstract}

\section{Introduction}
Neural networks are inherently susceptible to adversarial perturbations. Adversarial perturbations fool neural networks by adding an imperceptible amount of noise which leads to an incorrect prediction with high confidence \citep{carlini_adversarial_2017,goodfellow_explaining_2014,kurakin_adversarial_2016,szegedy_intriguing_2013}. 
There has been a lot of work on building defenses against adversarial perturbations \citep{papernot_distillation_2015,kannan_adversarial_2018}; the most commonly used defense is adversarial training as proposed by \citet{madry_towards_2017} and its variants \citep{zhang_theoretically_2019,pang_boosting_2020,huang_self-adaptive_2020,rice_overfitting_2020, gowal_uncovering_2020}, which use adversarially perturbed images at each training step as training data. Earlier studies \citep{kurakin2016adversarial,xie_feature_2018} showed that using adversarial samples during training leads to performance degradation on clean images. However, AdvProp \citep{xie_adversarial_2019} challenged this observation by showing that adversarial training can act as a regularizer, and therefore improve nominal accuracy, when using dual batch normalization (BatchNorm) layers \citep{ioffe2015batch} to disentangle the clean and adversarial distributions. 

\begin{figure}[t]
\centering
\subfigure[ \emph{Clean mode}]{\includegraphics[width=0.3\linewidth,trim=0cm 0.5cm 0cm 0.5cm,clip]{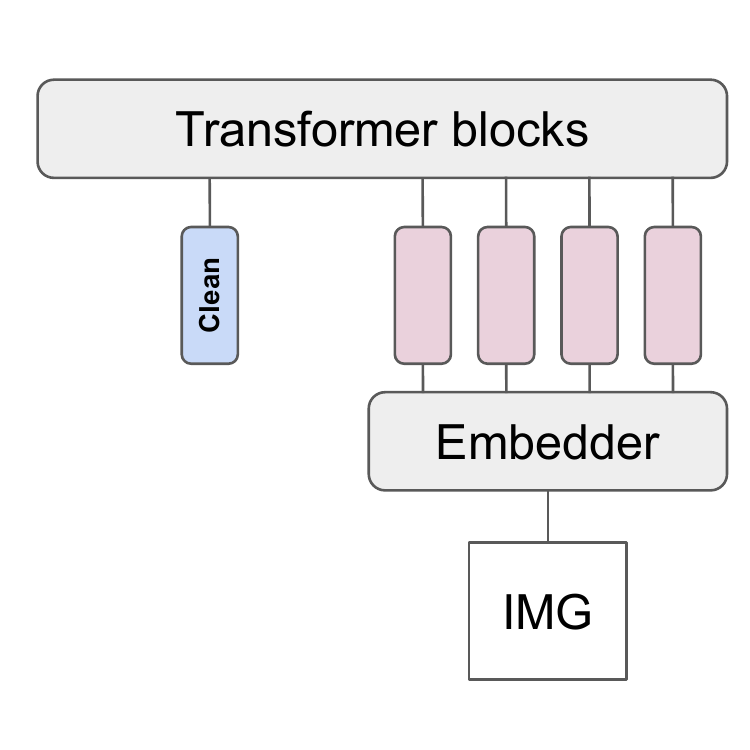}}
\hspace*{1\columnsep}
\subfigure[ \emph{Robust mode}]{\includegraphics[width=0.3\linewidth,trim=0cm 0.5cm 0cm 0.5cm,clip]{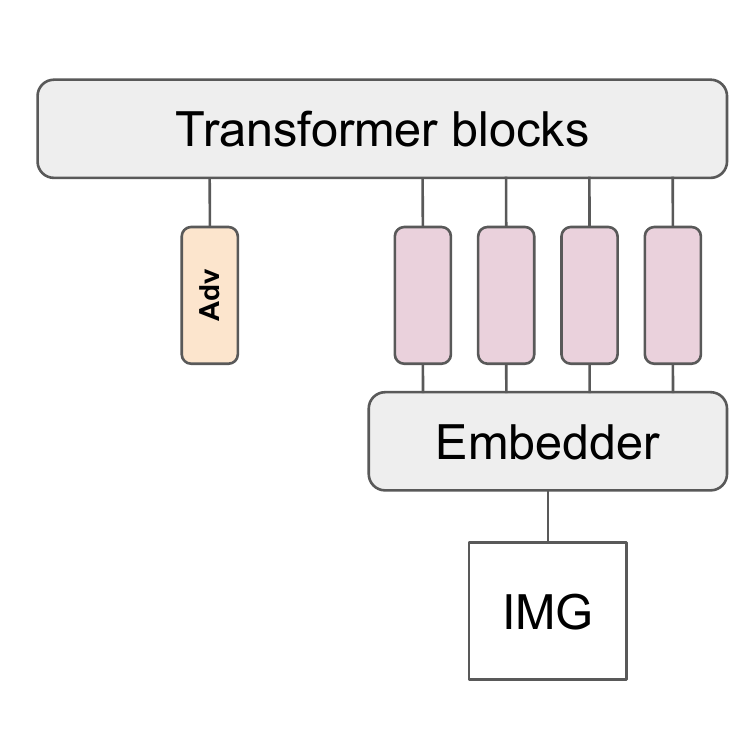}}
\hspace*{1\columnsep}
\subfigure[ \emph{Model soup}]{\includegraphics[width=0.3\linewidth,trim=0cm 0.5cm 0cm 0.5cm,clip]{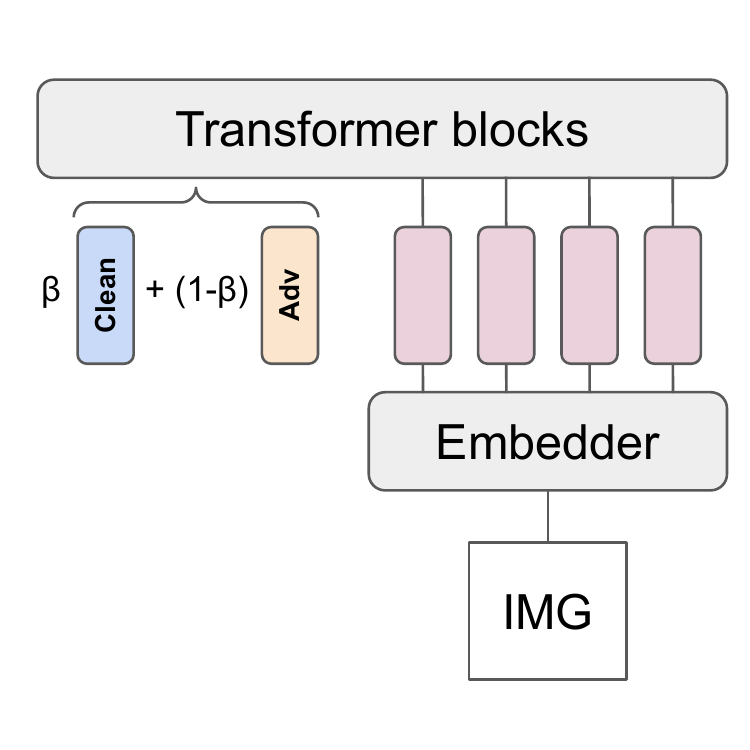}}
\caption{\textbf{Classification token as adapter.} The image is embedded into visual tokens (in pink). The behaviour of the model can set to the \emph{clean mode}, \emph{robust mode} or a \emph{model soup} by respectively using the \emph{clean} token (in blue), the \emph{adversarial} token (in orange) or a linear combination of these two tokens. The parameters of the embedder, the transformer blocks and the classifier are shared between modes.\label{fig:summary}}
\end{figure}

We draw attention to the broad similarity between the AdvProp approach and the adapters literature \citep{rebuffi2017learning, houlsby2019parameter} where a single backbone network is trained on multiple domains by means of adapters, where a few parameters specific to each domain  are trained separately while the rest of the parameters are shared. In light of this comparison, we further develop the line of work introduced by AdvProp and analyze it from an adapter perspective. In particular, we explore various adapters and aim to obtain the best classification performance with minimal additional parameters. Our contributions are as follows:%

\squishlist
    \item We show that, in order to benefit from co-training on clean and adversarial samples, it is not necessary to separate the batch statistics of clean and adversarial images in BatchNorm layers. We demonstrate empirically that it is enough to use domain specific trainable parameters 
    to achieve similar results.
    \item Inspired by the adapters literature, we evaluate various adapters. We show that training separate classification tokens of a \vit for the clean and adversarial domains is enough to match the classification performance of dual normalization layers with $49 \times$ fewer domain specific parameters. This classification token acts as a conditioning token which can modify the behaviour of the network to be either in \emph{clean} or \emph{robust mode} (Figure \ref{fig:summary}).
    \item Unlike \citet{xie_adversarial_2019} and \cite{herrmann2022pyramid}, we also aim at preserving the robust performance of the network against adversarial attacks. We show that our conditional token can obtain SOTA nominal accuracy in the \emph{clean mode} while at the same time achieving competitive \linf-robustness in the \emph{robust mode}. As a by-product of our study, we show that adversarial training of \vitb on \imagenet leads to state-of-the-art robustness against \linf-norm bounded perturbations of size $4/255$.
    \item We empirically demonstrate that training with adapters enables \emph{model soups} \citep{wortsman2022model}. This allow us to introduce \emph{adversarial model soups}, models that trade-off between clean and robust accuracy through linear interpolation of the clean and adversarial adapters. To the best of our knowledge, our work is the first to study \emph{adversarial model soups}. We also show that \emph{adversarial model soups} perform better on \imagenet variants than the state-of-the-art with masked auto-encoding~\citep{he2022masked}.
\squishend

\section{Related Work}
\paragraph{Adversarial training.}
Although more recent approaches have been proposed, the most successful method to reduce the vulnerability of image classifiers to adversarial attacks is \emph{adversarial training}, which generates on-the-fly adversarial counterparts for the training images and uses them to augment the training set \citep{croce2020robustbench}. \citet{goodfellow_explaining_2014} used the single-step Fast Gradient Sign Method (FGSM) attack to craft such adversarial images.
Later, \citet{madry_towards_2017} found that using iterative Projected Gradient Descent (\pgd{}) yields models robust to stronger attacks. Their scheme has been subsequently improved by several modifications, e.g.~a different loss function \citep{zhang_theoretically_2019}, unlabelled or synthetic data \citep{carmon_unlabeled_2019, uesato_are_2019, gowal2021improving}, model weight averaging \citep{gowal_uncovering_2020}, adversarial weight perturbations \citep{wu2020adversarial}, and better data augmentation \citep{rebuffi2021data}. While the main drawback of adversarial training is the degradation of performance of robust models on clean images \citep{tsipras_robustness_2018}, \citet{xie_adversarial_2019} showed that adversarial images can be leveraged as a strong  regularizer to {\em improve} the clean accuracy of classifiers on \imagenet. In particular, they propose AdvProp, which introduces separate BatchNorm layers specific to clean or adversarial inputs, with the remaining layers being shared. This approach and the role of normalization layers when training with both clean and adversarial points has been further studied by \citet{xie2019intriguing, walter2022fragile}. Recently, \citet{wang2022removing} suggest removing BatchNorm layers from the standard \resnet architecture \citep{he2015deep} to retain high clean accuracy with adversarial training, but this negatively affects the robustness against stronger attacks.\footnote{See {\tiny\url{https://github.com/amazon-research/normalizer-free-robust-training/issues/2}}.} Finally, \citet{kireev_effectiveness_2021, herrmann2022pyramid} showed that carefully tuning the threat model in adversarial training might improve the performance on clean images and in the presence of distribution shifts, such as \emph{common corruptions} \citep{hendrycks_benchmarking_2018}.

\paragraph{Adapters.} In early work on deep neural networks, \citet{caruana1997multitask} showed that sharing network parameters among tasks acts as a regularizer. Aiming at a more efficient parameter sharing, \citet{rebuffi2017learning,rosenfeld2018incremental} introduced adapters -- small training modules specific to each task which can be stitched all along the network. In other lines of work, \citet{mallya2018piggyback,mancini2018adding} adapt a model to new tasks using efficient weight masking and \citet{li2016revisiting,maria2017autodial} perform domain adaptation by batch statistics modulation. While these approaches require having as many adapters as tasks, \citet{perez2018film} propose an adapter layer whose weights are generated by a conditioning network. Besides computer vision, adapters are also used in natural language processing for efficient fine-tuning \citep{houlsby2019parameter,pfeiffer2020adapterfusion,wang2020k} and multi-task learning \citep{stickland2019bert}.

\paragraph{
Merging multiple models.} While  ensembles are a popular and successful way to combine multiple independently trained classifiers to improve on individual performance \citep{ovadia2019can,gontijo2021no}, they increase the inference cost as they require a forward pass for each sub-network of the ensemble. An alternative approach is taken by \citet{wortsman2022model} who propose to fine-tune a fully trained model with different hyperparameter configurations and then average the entire set of weights of the various networks. The obtained \emph{model soups} get better performance than each individual model and even ensembles.
Model soups are in spirit similar to Stochastic Weight Averaging \citep{izmailov_averaging_2018} which consists in averaging weights along an optimization trajectory rather than averaging over independent runs.

\section{Method}

\subsection{Co-training with nominal and adversarial training}

\citet{goodfellow_explaining_2014} propose adversarial training as a way to regularize standard training. They jointly optimize the model parameters $\vtheta$ on clean and adversarial images 
with the co-training loss
\begin{equation}
 \alpha L(f(\vx; \vtheta), y) + (1 - \alpha) \maximize_{\vdelta \in \sS} L(f(\vx + \vdelta; \vtheta), y),
\label{eq:co_training_loss}
\end{equation}
\noindent where pairs of associated examples $\vx$ and labels $y$ are sampled from the training dataset, $f(\cdot; \vtheta)$ is a model parametrized by $\vtheta$, $L$ defines the loss function (such as the cross-entropy loss in the classification context), and $\sS$ is the set of allowed perturbations. 
Setting $\alpha=1$ boils down to nominal training on clean images and setting $\alpha=0$ leads to adversarial training as defined by \citet{madry_towards_2017}. In our case, we consider $\ell_\infty$ norm-bounded perturbations of size $\epsilon=4/255$, so we have $\sS = \{ \vdelta ~|~ \| \vdelta \|_\infty \leq \epsilon \}$, and we use untargeted attacks to generate the adversarial perturbations $\vdelta$ (see details in Section~\ref{sec:experiments}).

\subsection{Separating batch statistics is not necessary}\label{subsec:batch_stats}
BatchNorm is a widely used normalization layer shown to improve performance and training stability of image classifiers \citep{ioffe2015batch}. We recall that a BatchNorm layer, given a batch as input, first normalizes it by subtracting the mean and dividing by the standard deviation computed over the entire batch, then it applies  an affine transformation, with learnable scale and offset parameters. During training, it accumulates these so-called \emph{batch statistics} to use during test time, so that the output of the classifier for each image is independent of the other images in the batch. The batch statistics can be seen an approximation of the statistics over the image distribution.

\citet{xie_adversarial_2019} show that optimizing the co-training loss in Eq.~\ref{eq:co_training_loss} can yield worse results on clean images than simple nominal training. This is especially the case when the network has a low capacity or the attack (i.e., the inner maximization) is too strong (such as using a large perturbation radius $\epsilon$). To solve this issue, they propose AdvProp, which consists in using distinct BatchNorm layers for clean and adversarial images. They argue that {\it ``maintaining one set of [BatchNorm] statistics results in incorrect statistics estimation''}, which could be the reason for the performance degradation.
%
We note that using two sets of BatchNorm layers for the clean and adversarial samples as in AdvProp creates two sets of batch statistics but also two sets of learnable scale and offset parameters.
In the following we investigate whether having separate batch statistics is a necessary condition for successful co-training.

\begin{figure}
  \centering
  \includegraphics[width=\linewidth]{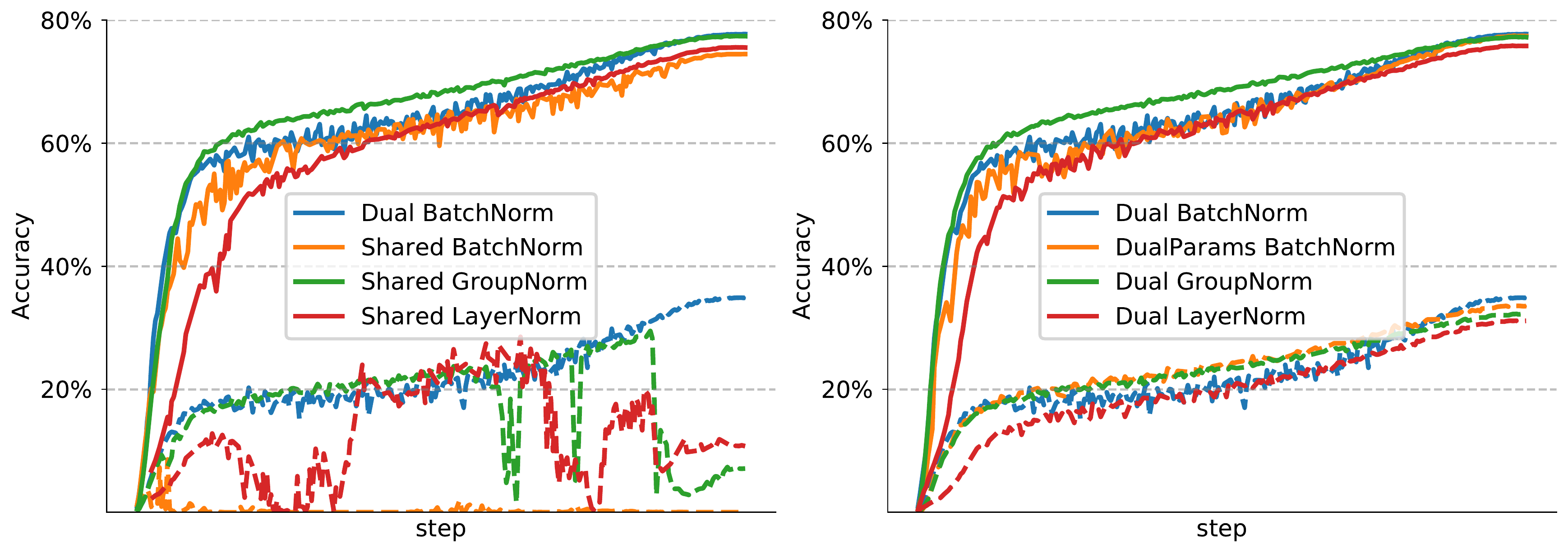}
  \caption{{\bf Dual parameters are enough.} We report the clean (solid lines) and robust accuracy (dashed lines) over training steps of \resnet-50 trained on \imagenet with the co-training loss of Eq.~\ref{eq:co_training_loss} ($\epsilon=4/255$): for models with dual layers. Clean accuracy refers to the \emph{clean mode} and the robust accuracy to the \emph{robust mode}. \textit{Left panel.} 
  We compare models with different normalization layers with no domain-specific parameters (Shared BatchNorm, Shared LayerNorm, Shared GroupNorm) to Dual BatchNorm as proposed by \citet{xie_adversarial_2019}: regardless of the type of normalization, the robustness of classifiers without dual layers drops to (almost) zero at the end of training. 
  \textit{Right panel.} We use domain-specific normalization layers (Dual BatchNorm, Dual LayerNorm, Dual GroupNorm) and a model with BatchNorm with shared batch statistics but domain-specific scale and offset (DualParams BatchNorm): all models achieve high clean and robust accuracy. \label{fig:curves_rn50}}%
\end{figure}

Figure~\ref{fig:curves_rn50} shows the clean and robust accuracy of various model architectures as training progresses. The left panel demonstrates that, if we share both batch statistics and scales/offsets (Shared BatchNorm, orange curves), the robust accuracy (orange dashed line) quickly drops, far from the one obtained by AdvProp (Dual BatchNorm, blue curve) which is above $34\%$. However, if we use a single set of batch statistics but specific scales and offsets for clean and adversarial images, we can observe on the right panel of Figure \ref{fig:curves_rn50} that the robust accuracy (DualParams BatchNorm, orange dashed line) matches the one (blue dashed line) obtained by AdvProp. This demonstrates that it is possible to achieve nominal and robust classification results similar to those of AdvProp without separate batch statistics.

Furthermore, there exist normalization layers such as LayerNorm \citep{ba2016layer} or GroupNorm \citep{wu2018group} which do not use batch statistics, as their normalization step is done per sample and not per batch. Hence, according to the hypothesis of \citet{xie_adversarial_2019}, these types of normalization layer should not suffer from performance degradation. Nevertheless, the left panel of Figure \ref{fig:curves_rn50} shows that their robust accuracy (green and red dashed lines) does not match the robust accuracy of AdvProp (Dual BatchNorm), and is unstable over training steps. However, by making the scales and offsets of LayerNorm and GroupNorm specific to clean and adversarial images, 
their robust accuracy matches that obtained with dual BatchNorm layers, as shown in the right panel of Figure~\ref{fig:curves_rn50}. This suggests that a key element to make the co-training loss of Eq.~\ref{eq:co_training_loss} work for various normalization layers is to have \uline{trainable parameters which are specific to the clean and adversarial images}.\footnote{Interestingly, contrary to our observation that standard GroupNorm fails to retain robustness, \cite{xie2019intriguing} report that GroupNorm matches Dual BatchNorm. We explain this difference as we use a stronger untargeted attack in this manuscript compared to the targeted attack of \cite{xie2019intriguing}. Using a stronger attack allows us to reveal failure modes that would have been 
hidden otherwise.}

\subsection{Revisiting adapters with adversarial training}
The last observation strongly relates this setting to the adapters literature where a single backbone architecture has some parameters, called adapters, which are specific to different domains while the rest of the parameters are shared among tasks. In our case, the clean images form one domain and the adversarial images constitute another domain. In this work, we go beyond having separate normalization layers for the clean and adversarial images and investigate other types of adapters.

Formally, the model parameters $\vtheta$ can be decomposed into parameters $\vpsi$ which are shared among domains and parameters $\vphi$ which are specific to a domain. We call $\vphi_\textrm{clean}$ the parameters used when training on clean images and $\vphi_\textrm{adv}$ the parameters used when training on adversarial images. For example, in Section~\ref{subsec:batch_stats}, when we used dual LayerNorm layers, the scales and offsets of these normalization layers are contained in $\vphi_\textrm{clean}$ and $\vphi_\textrm{adv}$ whereas the rest of the model parameters are in $\vpsi$. Based on Eq.~\ref{eq:co_training_loss}, we optimize the following loss:
\begin{equation}
 \alpha L(f(\vx; \vpsi \cup \vphi_\textrm{clean}), y) + (1 - \alpha) \maximize_{\vdelta \in \sS} L(f(\vx + \vdelta; \vpsi \cup \vphi_\textrm{adv}), y).
\label{eq:adapters_loss}
\end{equation}
Finally, we introduce some notation for models with adapters at inference time: we call $f(\cdot; \vpsi \cup \vphi_\textrm{clean})$ the \emph{clean mode} for prediction as we use the adapters $\vphi_\textrm{clean}$ trained on the clean data. Conversely, we call $f(\cdot; \vpsi \cup \vphi_\textrm{adv})$ the \emph{robust mode} when using the adapters $\vphi_\textrm{adv}$ trained on the perturbed data.

\subsection{Training with adapters enables adversarial model soups}
\citet{wortsman2022model} propose \emph{model soups}, which consist in averaging the weights of multiple models fine-tuned from the same pre-trained model. The resulting weight averaged model can benefit from the original models without incurring any extra compute and memory cost at inference time. Currently, in our setting the user would have to know at test time if the network should be in \emph{clean} or \emph{robust mode}. A \emph{model soup}, by its ability to merge models, is a way to bypass this issue. We formulate the hypothesis that \uline{training with adapters enables \emph{model soups}}. With this in mind, we observe that training with adapters means that most of the model parameters are already shared, so \emph{model souping} would simply consist in linearly interpolating the weights of the adapters for the two modes. We call \emph{adversarial model soups}, the model soups with a model co-trained on clean and adversarial samples.
We get the following parametrized model:
\begin{equation}
 f(\cdot; \vpsi \cup (\beta \vphi_\textrm{clean} + (1-\beta)\vphi_\textrm{adv}))
\label{eq:model_soup}
\end{equation}
where $\beta$ is the weighting factor when averaging the adapters. If $\beta=1$, the \emph{model soup} boils down to the \emph{clean mode} and conversely $\beta=0$ corresponds to the \emph{robust mode}. In Section \ref{subsec:model_soups}, we assess this hypothesis and show that forming \emph{model soups} between independent nominal and robust models fails.

\section{Experimental setup} \label{sec:experiments}

\paragraph{Architecture.} We focus our study on the B16 variant of the Vision Transformer (\vitb) introduced by \cite{dosovitskiy2020image}. We adopt the modifications proposed by \cite{he2022masked}: the linear classifier is applied on the mean of the final tokens except the classification token. We train this network by using supervised training from scratch as proposed in \cite{he2022masked} (see the appendix).

\paragraph{Attacks.}
We consider adversarial robustness against untargeted \linf-bounded attacks with radius $\epsilon=4/255$. This is the most common setup for \imagenet models, and it is more challenging to defend against than the targeted threat model used by \citet{xie2019intriguing}. To generate the adversarial perturbations we use Projected Gradient Descent \citep{madry_towards_2017} with 2 steps named \pgd{2} (see details in the appendix) at training time and with 40 steps for evaluation (\pgd{40}).

\paragraph{Datasets.}
We focus our experimental evaluation on the \imagenet dataset \citep{russakovsky2015imagenet}, with images at $224\times 224$ resolution for both training and testing, as this is the standard large-scale benchmark for SOTA models and was used by \citet{xie_adversarial_2019} for AdvProp. We report clean and adversarial accuracy on the whole validation set. Moreover, we test the robustness against distribution shifts via several \imagenet variants: \imagenetc \citep{hendrycks_benchmarking_2018}, \imageneta \citep{hendrycks_natural_2019}, \imagenetr \citep{hendrycks_many_2020}, \imagenetsketch \citep{wang2019learning}, and Conflict Stimuli \citep{geirhos_imagenet-trained_2018}.

\section{Experimental results}

Similarly to our observation in Section \ref{subsec:batch_stats} for a \resnet-50, a fully shared \vitb trained with the co-training loss Eq. \ref{eq:co_training_loss} fails to retain any robustness. Therefore, we first investigate various adapters for \vitb to find an efficient training setting in Section \ref{subsec:efficient_adapter}. Then we study \emph{adversarial model soups} with adapters in Section \ref{subsec:model_soups} and finally show that training with adapters generalizes to other datasets and threat models.

\subsection{Finding an efficient setting} \label{subsec:efficient_adapter}

\paragraph{Choice of adapter.} Using adapters increases the number of parameters as the layers which we choose as adapters have twice as many parameters: one set of parameters for clean images and another for adversarial images. Hence, to avoid increasing the network memory footprint too heavily, we restrict our adapters study to layers with few parameters, thus excluding self-attention 
\citep{vaswani2017attention} layers and MLP layers. This leaves the options of having dual embedder, positional embedding, normalization layers or classification token; among them, the classification token has by far the least amount of parameters, 49-770$\times$ fewer than the other candidates (see details in Table~\ref{tab:adapter_type}). We must still verify that so few parameters are enough to preserve the advantages of the AdvProp training scheme. Hence, we train a model for each type of adapter and
compare them with two models without adapters, one trained with nominal training and the other with adversarial training. We observe in Table \ref{tab:adapter_type} that by using two classification tokens as adapters, which means only 768 extra parameters out of 86M, we reach 83.56\% clean accuracy on \imagenet, which is an improvement of +0.92\% over standard training. Moreover, we obtain a robust accuracy of 49.87\% in the \emph{robust mode}, which is close to the robust accuracy given by adversarial training. 
Notably, we see that adapting other layers with more parameters such as all LayerNorm scales and offsets results in similar performances in both \emph{clean} and \emph{robust modes}. 
This indicates that 
 {\it (i)} it is not necessary to split the normalization layers to reproduce the effect of AdvProp,
 and {\it (ii)} even a very small amount of dual parameters provide sufficient expressiveness to adapt the shared portion of the network to the two modes.
Therefore, in the rest of the manuscript we focus on dual classification tokens as it requires the smallest number of extra parameters.

\paragraph{Number of attack steps.} As the results in Table \ref{tab:adapter_type} were obtained with \pgd{2}, we check if we can reduce the number of attack steps to be more computationally efficient. In Table \ref{tab:number_steps}, we report the results for two one-step methods: \textsc{N-FGSM} by \citet{de2022make} and \textsc{Fast-AT} by \cite{wong2020fast}. These methods perform more than 1\% worse in clean accuracy than nominal training while having no robustness at all in the \emph{robust mode}. 
We hypothesize that this is due to \emph{catastrophic overfitting} \citep{wong2020fast}: the adversarial perturbations generated with single-step attacks are not able to fool the classifier (\emph{robust} branch), which in turn cancels their positive effect on the \emph{clean} branch.
We also increase the number of attack steps to 5 with \pgd{5}. We notice a small improvement over \pgd{2} of 0.4\% in robust accuracy while the clean accuracy is on par. Hence, \pgd{2} seems to be a good compromise between efficiency and classification performance.

\begin{table}[ht]
\begin{floatrow}
\capbtabbox{%
\resizebox{0.6\textwidth}{!}{%
  \begin{tabular}{l|c|cc|cc}
    \hline
    \cellcolor{header} \textsc{Setup}  & \cellcolor{header} \textsc{Adapter} & \multicolumn{2}{c}{\cellcolor{header} \textsc{Clean Mode}} & \multicolumn{2}{c}{\cellcolor{header} \textsc{Robust Mode}}  \Tstrut \\
    \cellcolor{header}  & \cellcolor{header} \# params & \cellcolor{header} \textsc{Clean} & \cellcolor{header} \textsc{Robust} & \cellcolor{header} \textsc{Clean} & \cellcolor{header} \textsc{Robust} \Tstrut \\
    \hline
    \hline
    \multicolumn{6}{l}{\cellcolor{subheader} \textsc{Baselines}} \TBstrut \\
    \hline
    Nominal Training  &  0 & 82.64\% & 0\% & - & - \Tstrut \\
    Adversarial Training &  0 & 76.88\% & 56.19\% & - & - \\
    Co-training (Eq. \ref{eq:co_training_loss}) &  0 & 82.37\% & 0\% & - & - \Bstrut\\
    \hline
    \hline
    \multicolumn{6}{l}{\cellcolor{subheader} \textsc{With adapters}} \TBstrut \\
    \hline
    Embedder  &  591k & 83.51\% & 0.01\% & 77.68\%	& 50.69\% \Tstrut \\
    Positional embedding  &  151k & 83.65\% & 0.02\% & 77.75\% & 49.99\% \\
    All LN's scales/offsets &  38k & 83.63\% &	0\% & 77.65\% & 50.02\% \\
    Classification token  &  \textbf{0.8k} & 83.56\% & 0.01\% & 77.69\% & 49.87\%  \Bstrut \\
    \hline
    \end{tabular}
}
}{%
  \caption{{\bf Influence of the adapter type.} We report the clean and robust accuracies in \emph{clean} and \emph{robust mode} on \imagenet of networks trained with different layer types as adapters. We compare them with networks without adapters trained with nominal training and adversarial training. We recall that \vitb has 86M parameters. \label{tab:adapter_type}}%
}
\capbtabbox{%
\resizebox{0.36\textwidth}{!}{%
\begin{tabular}{l|c|cc}
    \hline
    \cellcolor{header} \textsc{Attack}  & \cellcolor{header} \# steps & \cellcolor{header} \textsc{Clean} & \cellcolor{header} \textsc{Robust}  \TBstrut \\
    \hline
    \textsc{Fast-AT} & 1 & 81.31\% & 0.00\%  \Tstrut \\
    \textsc{N-FGSM} & 1 & 81.00\% & 0.00\%   \\
    \pgd{2} & 2 & 83.56\% & 49.87\%   \\
    \pgd{5} & 5 & 83.60\% & 50.27\%  \Bstrut \\
    \hline
    \end{tabular}
}
}{%
  \caption{{\bf Influence of the number of attack steps.} We report the clean (in \emph{clean} mode) and robust accuracy (in \emph{robust} mode) on \imagenet 
  training with various number of attack steps. \label{tab:number_steps}}%
}
\end{floatrow}
\end{table}

\paragraph{Weighting the co-training loss.} In the co-training loss Eq. \ref{eq:co_training_loss}, the $\alpha$ hyperparameter controls how much the loss is weighted towards clean or adversarial samples. For example, setting $\alpha=0$ means we train solely on adversarial samples. In Figure \ref{fig:alphas}, where we evaluate several values for $\alpha$ (dividing the range between 0 and 1 into intervals of length $0.1$), we notice that 
only the values between $\alpha=0$ and $\alpha=0.4$ form a Pareto front that strictly dominates the other intervals. Indeed, between $\alpha=1$ and $\alpha=0.4$, decreasing $\alpha$ leads to better performance both in \emph{clean} and \emph{robust modes}. In fact, setting $\alpha=0.4$ leads to 83.76\% clean accuracy (in \emph{clean mode}) and 52.19\% robust accuracy (in \emph{robust mode}) which are both better than the values obtained in Table \ref{tab:adapter_type} with $\alpha=0.5$. In Figure \ref{fig:filters} (in the appendix), we visualize the filters of the embedder when training with various values of $\alpha$. We observe that for $\alpha=0.2$ and for $\alpha=0.8$ the filters look quite similar to the filters learned with adversarial training ($\alpha=0$) and nominal training ($\alpha=1$), respectively. Interestingly, filters learned with $\alpha=0.4$ and $\alpha=0.6$ are not the simple combination of nominal and adversarial filters but rather new visually distinct filters. This indicates that co-training on clean and adversarial samples can lead to a new hybrid representation for the shared layers compared to nominal and adversarial training.

\paragraph{Robustness to stronger attacks.} For completeness we further test the robustness of a subset of our models with a mixture of \autoattack~\citep{croce_reliable_2020} and \multitargeted~\citep{gowal_alternative_2019}, denoted by \textsc{AA+MT}. Pure adversarial training, which obtains 56.19\% robust accuracy against \pgd{40} (Table \ref{tab:adapter_type}), reaches 54.15\% robust accuracy against \textsc{AA+MT}. This is a new state-of-the-art robust accuracy on \imagenet, improving by +6.55\% over the 47.60\% reported by \cite{debenedetti2022light}. 
While \citet{debenedetti2022light} advocate for weak data augmentation for training robust \vit, our training procedure follows \citet{he2022masked} and contains heavy augmentations (see appendix): we conclude that large models still benefit from strong data augmentations even with adversarial training.
Finally, the \emph{robust mode} of the model co-trained with $\alpha=0.4$ in the previous paragraph reaches 49.55\% robust accuracy against \textsc{AA+MT}, which still surpasses the prior art and preserves competitive robust performance.


\subsection{Evaluating model soups} \label{subsec:model_soups}

\paragraph{Adapters enable \emph{adversarial model soups}.} One downside of using adapters is that one needs to know if for an input image the network should be put in \emph{clean} or \emph{robust mode}. This motivates \emph{adversarial model soups} which allow to create a single model performing well both in clean and robust accuracy. First, if we independently train two \vitb, one nominally and the other adversarially, and then try to perform \emph{model soups} on them, we notice in Table \ref{tab:indep_soup} (in the appendix) that both robust and clean accuracies drop immediately when the weighting factor $\beta$ between parameters is not equal to 0 or 1. 
We evaluate various \emph{model soups} with the models of Table \ref{tab:adapter_type}, meaning that the parameters specific to the clean and robust domain are averaged with weight $\beta$ to obtain a single classifier.
We notice in Figure \ref{fig:soup_adapters} (in the appendix) that \emph{adversarial model soups} work equally well with the various types of adapters, where sliding the value of $\beta$ allows to smoothly trade-off clean accuracy for robustness. 
This validates our hypothesis that adapters enable \emph{model soups}. 

\begin{figure}
\begin{floatrow}
\ffigbox{%
  \includegraphics[width=\linewidth]{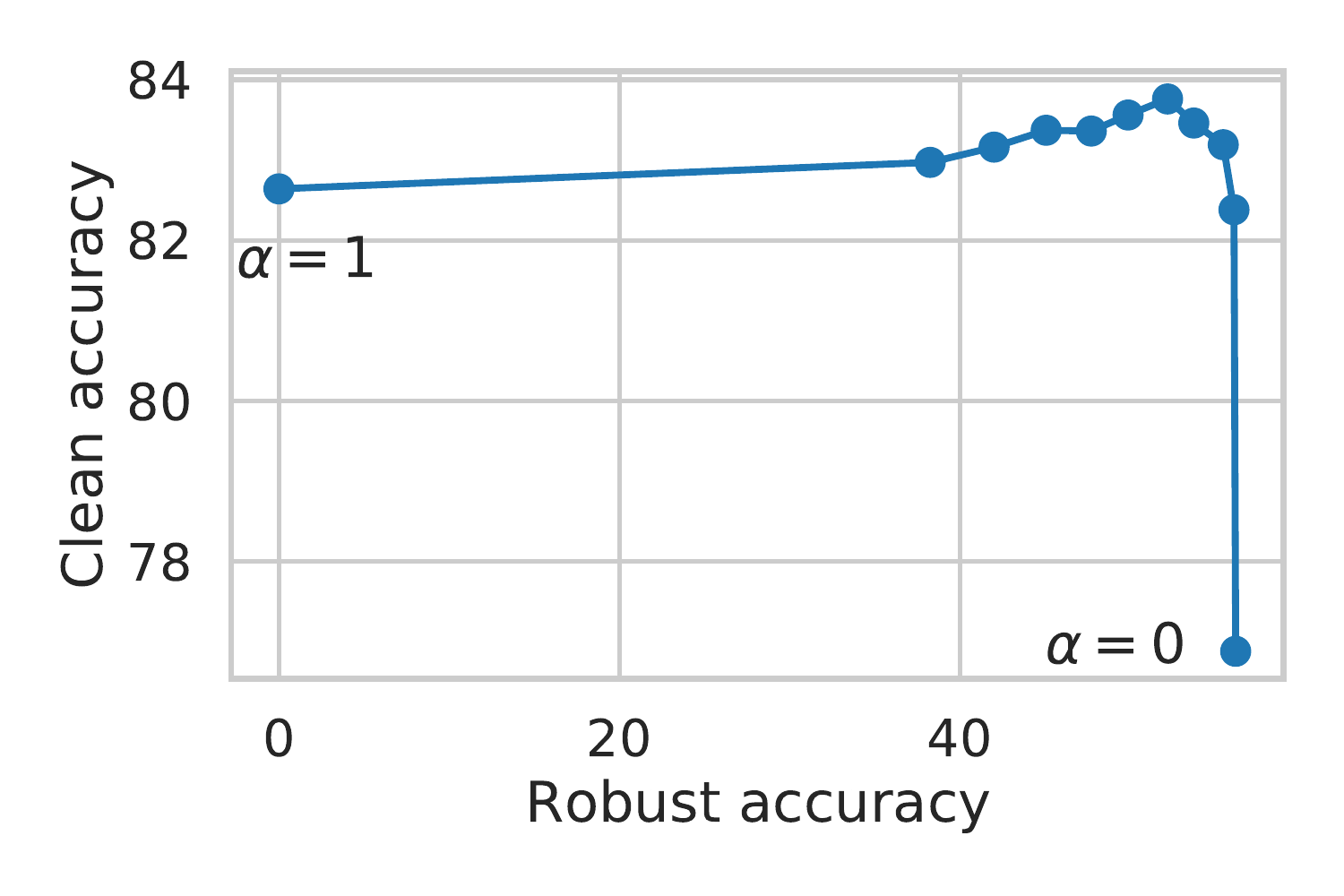}
}{%
  \caption{{\bf Co-training loss weighting.} We report the clean accuracy (in \emph{clean} mode) and robust accuracy (in \emph{robust} mode) on \imagenet when training with various weightings of the co-training loss Eq.~\ref{eq:adapters_loss}. We recall that $\alpha=1$ corresponds to pure nominal training and $\alpha=0$ to adversarial training. \label{fig:alphas}}%
}
\ffigbox{%
  \centering
  \includegraphics[width=\linewidth]{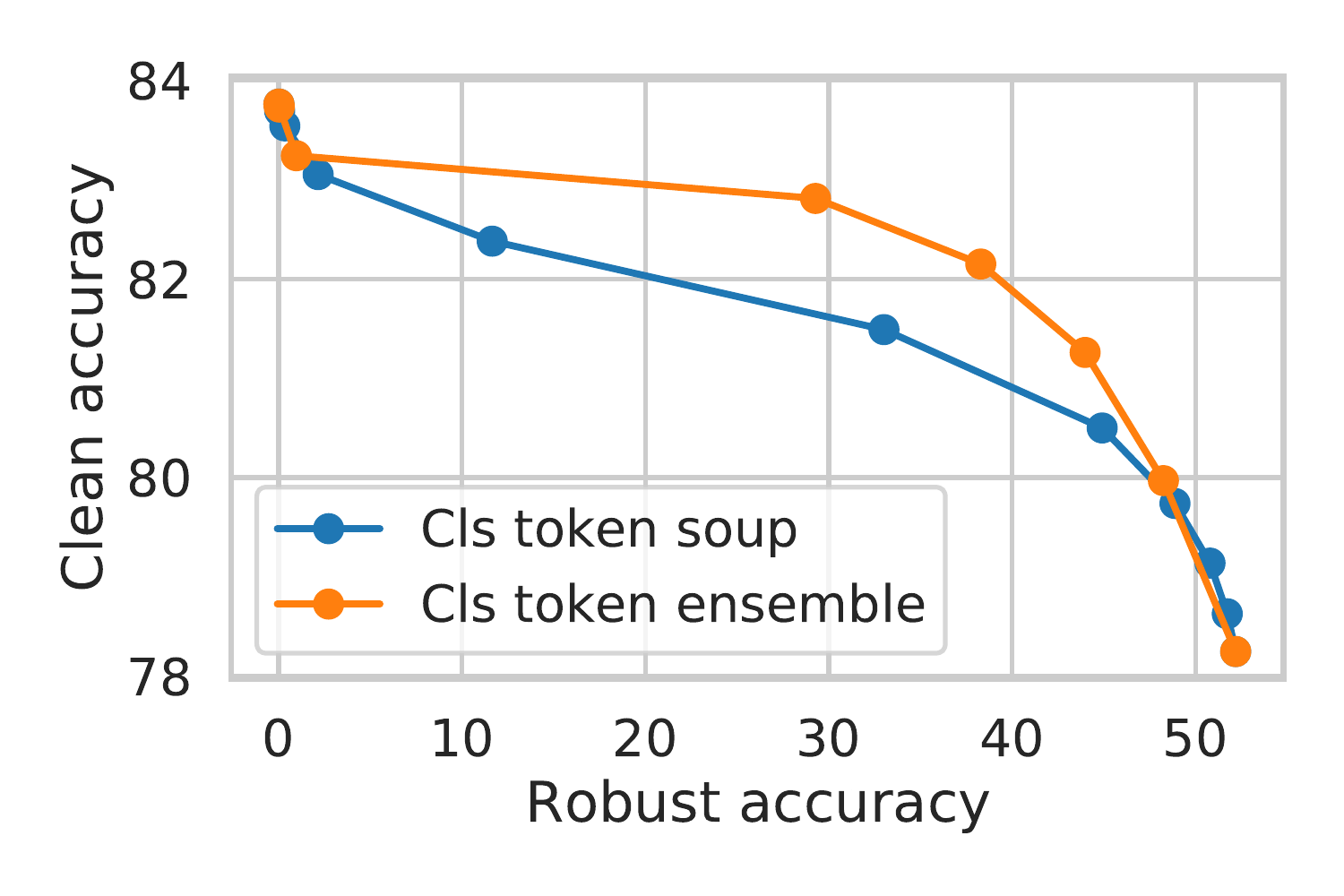}
}{%
  \caption{{\bf Soup vs ensemble.} We report the clean and robust accuracy (single mode for both) on \imagenet for various \emph{model soups} and ensembles of a same network trained with classification token adapters. \label{fig:soup_vs_ensemble}}%
}
\end{floatrow}

\end{figure}

\paragraph{Soup or ensemble.} In Figure \ref{fig:soup_vs_ensemble} we compare the classification 
performance of \emph{adversarial model soups} and ensembles obtained by linear combination of the \emph{clean} and \emph{robust modes} at the probability prediction level. We notice that ensembling produces a better Pareto front than \emph{adversarial model soup} but ensembles, with their two forward passes, require twice the compute of \emph{model soups}. Hence, \emph{adversarial model soups} allow to choose the trade-off between clean and robust accuracy with performance close to ensembling while only requiring the same compute as a single network.  

\paragraph{Extrapolation.} For the anecdote, we experiment with \emph{adversarial model soups} for extrapolation with values of the weighting factor $\beta$ above 1 and below 0. Interestingly, we observe that setting $\beta=1.05$ leads to 83.81\% clean accuracy which is better than the 83.76\% obtained in the \emph{clean mode}. Similarly, setting $\beta=-0.05$ leads to 52.26\% robust accuracy which is slightly better than the 52.19\% obtained in the \emph{robust mode}. Hence, it appears that \emph{adversarial model soups} do not need to be restricted to interpolation.

\begin{figure}
  \centering
  \includegraphics[width=0.9\linewidth]{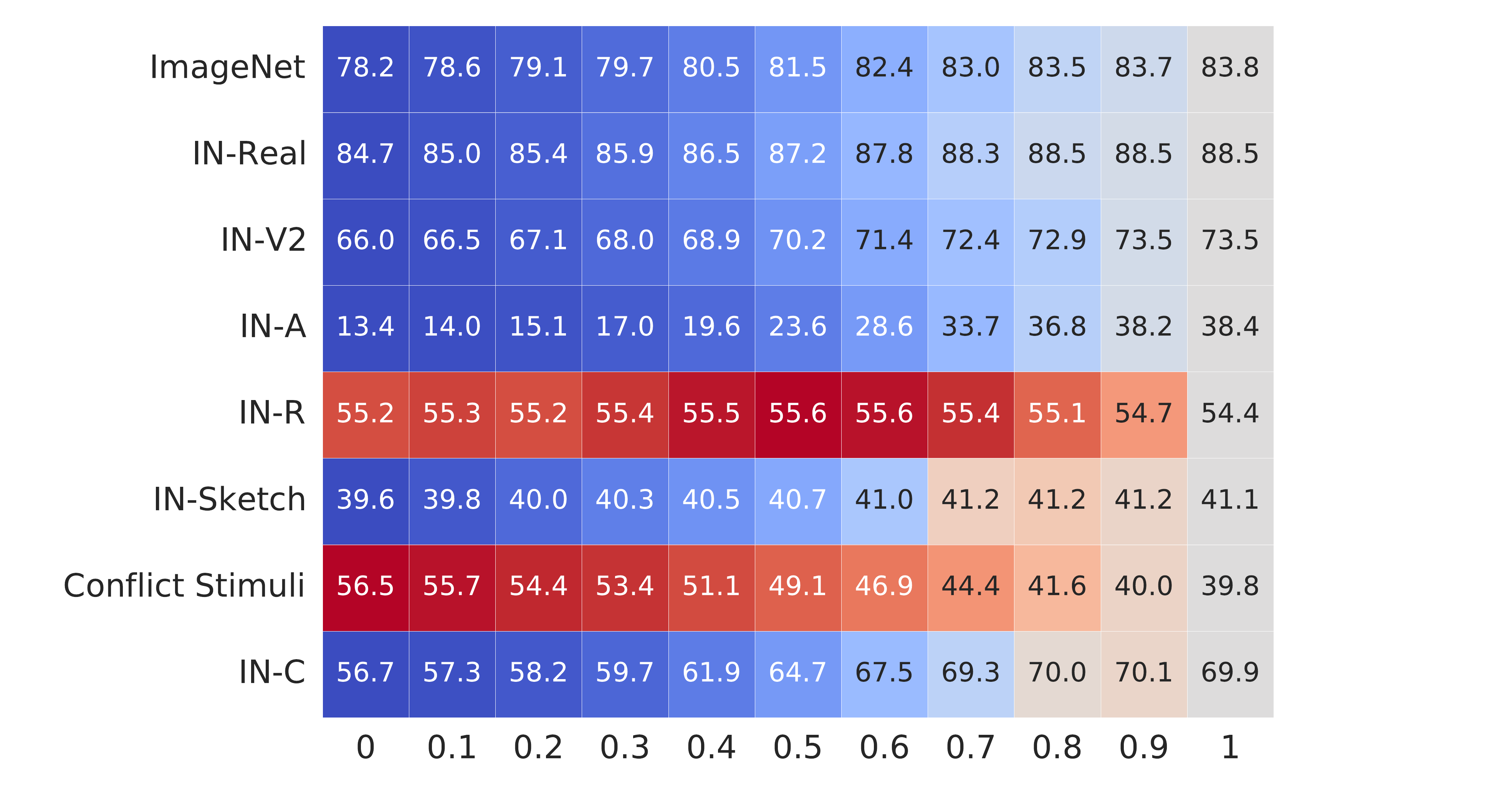}
  \caption{{\bf 
  Soups for \imagenet variants.} We report the accuracy on \imagenet variants for \emph{adversarial model soups}. The x-axis corresponds to the interpolation weighting factor $\beta$ where $\beta=1$ boils down to the \emph{clean mode} and $\beta=0$ to the \emph{robust mode}. Red (blue) means better (worse) than the \emph{clean mode} (in grey). 
  \label{fig:palette_soup}}%
\end{figure}

\paragraph{Soups for \imagenet variants.} As \emph{adversarial model soups} allow to create models with chosen trade-off between clean and robust accuracy, we might expect that such models perform better than nominal ones when distribution shifts occur. For example, \citet{kireev_effectiveness_2021} showed that adversarial training can even help with common corruptions when specifically tuned for such task (note that they use smaller datasets than \imagenet). 
We then compute the accuracy of \emph{adversarial model soups} with varying $\beta$ on \imagenet variants (results in Figure \ref{fig:palette_soup}):
while half of the best performance are obtained with the \emph{clean} classification token, for some variants such as \imagenetr, \imagenetc and \imagenetsketch the best results are obtained with 
intermediate tokens. Hence, \emph{adversarial model soups} can be used to reach a compromise between \imagenet variants to get the best average performance. Here $\beta=0.9$ yields the best mean accuracy 61.23\%. In Table \ref{tab:soup_vs_baselines}, we notice that this \emph{adversarial model soup} improves the mean accuracy by +4.00\% over a fine-tuned Masked Autoencoder (MAE-B16) checkpoint from \cite{he2022masked} and by +2.37\% over Pyramid-AT from \cite{herrmann2022pyramid}. It also improves by +2.24\% over the best performing ensemble of two networks trained independently with nominal and adversarial training respectively.

\begin{table}[ht]
 \caption{{\bf \imagenet variants results.} We report the \emph{adversarial model soup} achieving the best average accuracy over the eight variants and we compare it against several baselines. All models are trained on \imagenet and use the same \vitb architecture. \label{tab:soup_vs_baselines}}%
\begin{center}
\resizebox{1\textwidth}{!}{
\begin{tabular}{l|c|cccccccc|c}
    \hline
    \cellcolor{header} \textsc{Setup} & \cellcolor{header} \textsc{Compute} & \cellcolor{header} \textsc{ImageNet}  & \cellcolor{header} \textsc{IN-Real} & \cellcolor{header} \textsc{IN-V2} & \cellcolor{header} \textsc{IN-A} & \cellcolor{header} \textsc{IN-R} & \cellcolor{header} \textsc{IN-Sketch} & \cellcolor{header} \textsc{Conflict} & \cellcolor{header} \textsc{IN-C} & \cellcolor{header} \textsc{Mean}  \TBstrut \\
    \hline
    \hline
    \multicolumn{11}{l}{\cellcolor{subheader} \textsc{Baselines}} \TBstrut \\
    \hline
    Nominal training & $\times 1$ & 82.64\% & 87.33\% & 71.42\% & 28.03\% &   47.94\% & 34.43\% & 30.47\% & 64.45\% & 55.84\%   \Tstrut \\
    Adversarial training & $\times 1$  & 76.88\% & 83.91\% & 64.81\% & 12.35\% & 55.76\% & 40.11\% & \textbf{59.45\%} & 55.44\% & 56.09\%   \\
    Fine-tuned MAE-B16 & $\times 1$  & 83.10\% & 88.02\% & 72.80\% & 37.92\% & 49.30\% & 35.69\% & 27.81\% & 63.23\% & 57.23\%   \\
    Pyramid-AT (fully shared network) & $\times 1$  & 83.14\% & 87.82\% & 72.53\% & 32.72\% & 51.78\% & 38.60\% & 37.27\% & 67.01\% & 58.86\%   \\
    Independent networks ensemble & $\times 2$  & 82.86\% & 87.78\% & 71.73\% & 25.99\% & 54.20\% & 37.33\% & 46.41\% & 65.61\% & 58.99\%  \Bstrut \\
    \hline
    \hline
    \multicolumn{11}{l}{\cellcolor{subheader} \textsc{With adapters}} \TBstrut \\
    \hline
    Classification token soup & $\times 1$   & \textbf{83.69\%} & 88.50\% & \textbf{73.48\%} & \textbf{38.23\%} & 54.74\% & 41.17\% & 40.00\% & \textbf{70.07\%} & 61.23\% \Tstrut \\
    Classification token ensemble & $\times 2$  & 83.62\% & \textbf{88.58\%} & 73.36\% & 35.05\% & \textbf{56.32\%} & \textbf{41.36\%} & 49.14\% & 70.04\% & \textbf{62.18\%} \Bstrut  \\
    \hline
    \end{tabular}
}
\end{center}
\end{table}

\begin{table}[ht]
 \caption{{\bf Comparing threat models.} We optimize the co-training loss with adversarial examples coming from various attacks. For each run, we select the \emph{adversarial model soup} achieving the best mean performance on \imagenet variants, which we report in the table. We set the co-training loss weighting factor $\alpha=0.5$.  \label{tab:threat_models}}%
\begin{center}
\resizebox{1\textwidth}{!}{
\begin{tabular}{l|cccccccc|c}
    \hline
    \cellcolor{header} \textsc{Attack} & \cellcolor{header} \textsc{ImageNet}  & \cellcolor{header} \textsc{IN-Real} & \cellcolor{header} \textsc{IN-V2} & \cellcolor{header} \textsc{IN-A} & \cellcolor{header} \textsc{IN-R} & \cellcolor{header} \textsc{IN-Sketch} & \cellcolor{header} \textsc{Conflict} & \cellcolor{header} \textsc{IN-C} & \cellcolor{header} \textsc{Mean}  \TBstrut \\
    \hline
    Nominal training & 82.64\% & 87.33\% & 71.42\% & 28.03\% &   47.94\% & 34.43\% & 30.47\% & 64.45\% & 55.84\%   \TBstrut \\
    \hline
    \linf (untargeted) & 83.36\% & 88.31\% & 72.73\% & 34.81\% & 54.15\% & 40.37\% & 41.25\% & 69.54\% & \textbf{60.56\%}  \Tstrut \\
    \ltwo (untargeted) & 83.35\% & 88.12\% & 72.82\% & 35.17\% & 52.59\% & 39.34\% & 38.75\% & 68.99\% & 59.89\%   \\
    \linf (targeted) & 83.21\% & 87.97\% & 72.95\% & 33.87\% & 52.90\% & 38.70\% & 39.06\% & 68.49\% & 59.64\%   \\
    Pyramid-\linf & 83.62\% & 88.20\% & 73.36\% & 35.15\% & 53.94\% & 39.40\% & 41.88\% & 68.39\% & 60.49\% \Bstrut  \\
    \hline
    \end{tabular}
}
\end{center}
\end{table}

\subsection{Evaluating on other threat models and datasets} \label{subsec:small_datasets}

\paragraph{Evaluating other threat models.} \imagenet variants are also a good benchmark to compare different types of adversarial attack 
to generate the perturbations for the co-training loss in Eq.~\ref{eq:adapters_loss}: untargeted \linf-bounded perturbations with budget $\epsilon=4/255$ (our standard setup), untargeted \ltwo-bounded with $\epsilon \in  \{1, 2, 4, 8\}$, targeted (random target class as in \citealp{xie_adversarial_2019}) \linf-bounded with $\epsilon \in \{4/255, 8/255, 12/255\}$, and the Pyramid attack proposed by \cite{herrmann2022pyramid}. In Table \ref{tab:threat_models}, we select the best \emph{adversarial model soups} after training with 
each method a \vitb with dual classification tokens, and report its results on all variants. We see that the clean accuracy on the \imagenet validation set improves in all cases compared to standard training. Moreover, although the best performing attack varies across variants, we notice that the untargeted \linf attack achieves the best average accuracy.

\paragraph{Evaluating on other datasets.} We further test the effect of using the co-training loss with the classification token as adapter on other datasets. In Table \ref{tab:extra_datasets}, we see that our training procedure provides a consistent performance boost in clean accuracy compared to nominal training on \mnist \citep{lecun2010mnist}, \cifar, \cifarh \citep{krizhevsky_cifar-10_2014}, \textsc{SVHN} \citep{netzer2011reading}, \textsc{SUN-397} \citep{xiao2010sun}, \textsc{RESISC-45} \citep{cheng2017remote} and \textsc{DMLab} \citep{beattie2016deepmind}. This shows that our method generalizes well across datasets and can help regularize Vision Transformers on these smaller datasets, where they are known to perform worse compared to CNNs without pre-training \citep{zhang2021aggregating}. In Appendix \ref{sec:transfer_learning}, we also demonstrate that models pre-trained with co-training on \imagenet yield significantly better classification results when fine-tuning nominally on small datasets compared to fine-tuning from nominally and adversarially pre-trained models.

\begin{table}[t]
 \caption{{\bf Evaluating on other datasets.} We report the clean accuracy (in the \emph{clean mode}) when training on various datasets with several perturbation radii. We compare it with nominal training on the same architecture. \label{tab:extra_datasets}}%
\begin{center}
\resizebox{0.75\textwidth}{!}{
\begin{tabular}{l|ccccccc}
    \hline
    \cellcolor{header} \textsc{Radius}  & \cellcolor{header} \textsc{Mnist} & \cellcolor{header} \textsc{CIFAR-10} & \cellcolor{header} \textsc{CIFAR-100} & \cellcolor{header} \textsc{SVHN} &
    \cellcolor{header} \textsc{SUN-397} &
    \cellcolor{header} \textsc{RESISC-45} &
    \cellcolor{header} \textsc{DMLab}\TBstrut \\
    \hline
    Nominal & 99.48\% & 94.63\% & 79.21\% & 97.88\% & 65.24\% & 94.03\%	& 55.66\% \TBstrut \\
    \hline
    1/255 & 99.56\% & 96.68\% & 80.88\% & \textbf{98.38\%} & 67.08\% & 95.09\% & \textbf{58.94\%} \Tstrut \\
    2/255 & 99.67\% & \textbf{97.07\%} & 81.02\% & 98.14\%  & 67.27\% & \textbf{95.20\%} & 57.99\% \\
    4/255 & 99.68\% & 97.00\% & 80.97\% & 97.83\% & \textbf{67.32\%} & 94.86\% & 57.81\% \\
    8/255 & \textbf{99.71\%} & 96.34\% & \textbf{81.27\%} & 96.18\% & 65.97\% & 94.12\% & 53.64\% \Bstrut \\
    \hline
    \end{tabular}
}
\end{center}
\end{table}

\section{Conclusion}

In this work we have shown that adapters with a few hundreds of domain specific parameters are sufficient to switch between models with radically different behaviors. In fact, just replacing the classification token of a \vit can turn a classifier with SOTA nominal accuracy and no adversarial robustness into another one with robust accuracy close to that achieved with standard adversarial training. Moreover, merging the adapters allows to smoothly transition between the two modes, finding classifiers (i.e. our \emph{adversarial model soups}) with better performance on distribution shifts. These observations open up new interesting directions for future work to explore how to take advantage of the regularizing effect of adversarial training and whether it is possible to combine via soups other types of models.

\clearpage

\bibliography{iclr2023_conference}

\begin{thebibliography}{76}
\providecommand{\natexlab}[1]{#1}
\providecommand{\url}[1]{\texttt{#1}}
\expandafter\ifx\csname urlstyle\endcsname\relax
  \providecommand{\doi}[1]{doi: #1}\else
  \providecommand{\doi}{doi: \begingroup \urlstyle{rm}\Url}\fi

\bibitem[Ba et~al.(2016)Ba, Kiros, and Hinton]{ba2016layer}
Jimmy~Lei Ba, Jamie~Ryan Kiros, and Geoffrey~E Hinton.
\newblock Layer normalization.
\newblock \emph{arXiv preprint arXiv:1607.06450}, 2016.

\bibitem[Beattie et~al.(2016)Beattie, Leibo, Teplyashin, Ward, Wainwright,
  K{\"u}ttler, Lefrancq, Green, Vald{\'e}s, Sadik, et~al.]{beattie2016deepmind}
Charles Beattie, Joel~Z Leibo, Denis Teplyashin, Tom Ward, Marcus Wainwright,
  Heinrich K{\"u}ttler, Andrew Lefrancq, Simon Green, V{\'\i}ctor Vald{\'e}s,
  Amir Sadik, et~al.
\newblock Deepmind lab.
\newblock \emph{arXiv preprint arXiv:1612.03801}, 2016.

\bibitem[Carlini \& Wagner(2017)Carlini and Wagner]{carlini_adversarial_2017}
Nicholas Carlini and David Wagner.
\newblock Adversarial examples are not easily detected: {Bypassing} ten
  detection methods.
\newblock In \emph{Proceedings of the 10th {ACM} {Workshop} on {Artificial}
  {Intelligence} and {Security}}, pp.\  3--14. ACM, 2017.

\bibitem[Carmon et~al.(2019)Carmon, Raghunathan, Schmidt, Duchi, and
  Liang]{carmon_unlabeled_2019}
Yair Carmon, Aditi Raghunathan, Ludwig Schmidt, John~C Duchi, and Percy~S
  Liang.
\newblock Unlabeled data improves adversarial robustness.
\newblock In \emph{NeurIPS}, 2019.

\bibitem[Caruana(1997)]{caruana1997multitask}
Rich Caruana.
\newblock Multitask learning.
\newblock \emph{Machine learning}, 28\penalty0 (1):\penalty0 41--75, 1997.

\bibitem[Cheng et~al.(2017)Cheng, Han, and Lu]{cheng2017remote}
Gong Cheng, Junwei Han, and Xiaoqiang Lu.
\newblock Remote sensing image scene classification: Benchmark and state of the
  art.
\newblock \emph{Proceedings of the IEEE}, 105\penalty0 (10):\penalty0
  1865--1883, 2017.

\bibitem[Croce \& Hein(2020)Croce and Hein]{croce_reliable_2020}
Francesco Croce and Matthias Hein.
\newblock Reliable evaluation of adversarial robustness with an ensemble of
  diverse parameter-free attacks.
\newblock \emph{arXiv preprint arXiv:2003.01690}, 2020.

\bibitem[Croce et~al.(2020)Croce, Andriushchenko, Sehwag, Flammarion, Chiang,
  Mittal, and Hein]{croce2020robustbench}
Francesco Croce, Maksym Andriushchenko, Vikash Sehwag, Nicolas Flammarion, Mung
  Chiang, Prateek Mittal, and Matthias Hein.
\newblock Robustbench: a standardized adversarial robustness benchmark.
\newblock \emph{arXiv preprint arXiv:2010.09670}, 2020.

\bibitem[Cubuk et~al.(2020)Cubuk, Zoph, Shlens, and Le]{cubuk2019randaugment}
Ekin~D. Cubuk, Barret Zoph, Jonathon Shlens, and Quoc~V. Le.
\newblock Randaugment: Practical automated data augmentation with a reduced
  search space.
\newblock \emph{CVPR}, 2020.

\bibitem[de~Jorge et~al.(2022)de~Jorge, Bibi, Volpi, Sanyal, Torr, Rogez, and
  Dokania]{de2022make}
Pau de~Jorge, Adel Bibi, Riccardo Volpi, Amartya Sanyal, Philip~HS Torr,
  Gr{\'e}gory Rogez, and Puneet~K Dokania.
\newblock Make some noise: Reliable and efficient single-step adversarial
  training.
\newblock \emph{arXiv preprint arXiv:2202.01181}, 2022.

\bibitem[Debenedetti et~al.(2022)Debenedetti, Sehwag, and
  Mittal]{debenedetti2022light}
Edoardo Debenedetti, Vikash Sehwag, and Prateek Mittal.
\newblock A light recipe to train robust vision transformers.
\newblock \emph{arXiv preprint arXiv:2209.07399}, 2022.

\bibitem[Dosovitskiy et~al.(2020)Dosovitskiy, Beyer, Kolesnikov, Weissenborn,
  Zhai, Unterthiner, Dehghani, Minderer, Heigold, Gelly,
  et~al.]{dosovitskiy2020image}
Alexey Dosovitskiy, Lucas Beyer, Alexander Kolesnikov, Dirk Weissenborn,
  Xiaohua Zhai, Thomas Unterthiner, Mostafa Dehghani, Matthias Minderer, Georg
  Heigold, Sylvain Gelly, et~al.
\newblock An image is worth 16x16 words: Transformers for image recognition at
  scale.
\newblock \emph{arXiv preprint arXiv:2010.11929}, 2020.

\bibitem[Geirhos et~al.(2018)Geirhos, Rubisch, Michaelis, Bethge, Wichmann, and
  Brendel]{geirhos_imagenet-trained_2018}
Robert Geirhos, Patricia Rubisch, Claudio Michaelis, Matthias Bethge, Felix~A
  Wichmann, and Wieland Brendel.
\newblock {ImageNet}-trained {CNNs} are biased towards texture; increasing
  shape bias improves accuracy and robustness.
\newblock In \emph{International {Conference} on {Learning} {Representations}},
  2018.
\newblock URL \url{https://openreview.net/pdf?id=Bygh9j09KX}.

\bibitem[Gontijo-Lopes et~al.(2021)Gontijo-Lopes, Dauphin, and
  Cubuk]{gontijo2021no}
Raphael Gontijo-Lopes, Yann Dauphin, and Ekin~D Cubuk.
\newblock No one representation to rule them all: Overlapping features of
  training methods.
\newblock \emph{arXiv preprint arXiv:2110.12899}, 2021.

\bibitem[Goodfellow et~al.(2015)Goodfellow, Shlens, and
  Szegedy]{goodfellow_explaining_2014}
Ian~J Goodfellow, Jonathon Shlens, and Christian Szegedy.
\newblock Explaining and harnessing adversarial examples.
\newblock \emph{ICLR}, 2015.

\bibitem[Gowal et~al.(2019)Gowal, Uesato, Qin, Huang, Mann, and
  Kohli]{gowal_alternative_2019}
Sven Gowal, Jonathan Uesato, Chongli Qin, Po-Sen Huang, Timothy Mann, and
  Pushmeet Kohli.
\newblock An {Alternative} {Surrogate} {Loss} for {PGD}-based {Adversarial}
  {Testing}.
\newblock \emph{arXiv preprint arXiv:1910.09338}, 2019.

\bibitem[Gowal et~al.(2020)Gowal, Qin, Uesato, Mann, and
  Kohli]{gowal_uncovering_2020}
Sven Gowal, Chongli Qin, Jonathan Uesato, Timothy Mann, and Pushmeet Kohli.
\newblock Uncovering the limits of adversarial training against norm-bounded
  adversarial examples.
\newblock \emph{arXiv preprint arXiv:2010.03593}, 2020.
\newblock URL \url{https://arxiv.org/pdf/2010.03593}.

\bibitem[Gowal et~al.(2021)Gowal, Rebuffi, Wiles, Stimberg, Calian, and
  Mann]{gowal2021improving}
Sven Gowal, Sylvestre-Alvise Rebuffi, Olivia Wiles, Florian Stimberg,
  Dan~Andrei Calian, and Timothy~A Mann.
\newblock Improving robustness using generated data.
\newblock \emph{NeurIPS}, 34:\penalty0 4218--4233, 2021.

\bibitem[Goyal et~al.(2017)Goyal, Dollár, Girshick, Noordhuis, Wesolowski,
  Kyrola, Tulloch, Jia, and He]{goyal2017accurate}
Priya Goyal, Piotr Dollár, Ross Girshick, Pieter Noordhuis, Lukasz Wesolowski,
  Aapo Kyrola, Andrew Tulloch, Yangqing Jia, and Kaiming He.
\newblock Accurate, large minibatch sgd: Training imagenet in 1 hour.
\newblock \emph{arXiv preprint arXiv:1706.02677}, 2017.

\bibitem[He et~al.(2016)He, Zhang, Ren, and Sun]{he2015deep}
Kaiming He, Xiangyu Zhang, Shaoqing Ren, and Jian Sun.
\newblock Deep residual learning for image recognition.
\newblock \emph{CVPR}, 2016.

\bibitem[He et~al.(2022)He, Chen, Xie, Li, Doll{\'a}r, and
  Girshick]{he2022masked}
Kaiming He, Xinlei Chen, Saining Xie, Yanghao Li, Piotr Doll{\'a}r, and Ross
  Girshick.
\newblock Masked autoencoders are scalable vision learners.
\newblock In \emph{Proceedings of the IEEE/CVF Conference on Computer Vision
  and Pattern Recognition}, pp.\  16000--16009, 2022.

\bibitem[Hendrycks \& Dietterich(2018)Hendrycks and
  Dietterich]{hendrycks_benchmarking_2018}
Dan Hendrycks and Thomas Dietterich.
\newblock Benchmarking {Neural} {Network} {Robustness} to {Common}
  {Corruptions} and {Perturbations}.
\newblock In \emph{International {Conference} on {Learning} {Representations}},
  2018.
\newblock URL \url{https://openreview.net/pdf?id=HJz6tiCqYm}.

\bibitem[Hendrycks et~al.(2019)Hendrycks, Zhao, Basart, Steinhardt, and
  Song]{hendrycks_natural_2019}
Dan Hendrycks, Kevin Zhao, Steven Basart, Jacob Steinhardt, and Dawn Song.
\newblock Natural adversarial examples.
\newblock \emph{arXiv preprint arXiv:1907.07174}, 2019.

\bibitem[Hendrycks et~al.(2020)Hendrycks, Basart, Mu, Kadavath, Wang, Dorundo,
  Desai, Zhu, Parajuli, Guo, Song, Steinhardt, and Gilmer]{hendrycks_many_2020}
Dan Hendrycks, Steven Basart, Norman Mu, Saurav Kadavath, Frank Wang, Evan
  Dorundo, Rahul Desai, Tyler Zhu, Samyak Parajuli, Mike Guo, Dawn Song, Jacob
  Steinhardt, and Justin Gilmer.
\newblock The {Many} {Faces} of {Robustness}: {A} {Critical} {Analysis} of
  {Out}-of-{Distribution} {Generalization}.
\newblock \emph{arXiv preprint arXiv:2006.16241}, 2020.
\newblock URL \url{https://arxiv.org/pdf/2006.16241}.

\bibitem[Herrmann et~al.(2022)Herrmann, Sargent, Jiang, Zabih, Chang, Liu,
  Krishnan, and Sun]{herrmann2022pyramid}
Charles Herrmann, Kyle Sargent, Lu~Jiang, Ramin Zabih, Huiwen Chang, Ce~Liu,
  Dilip Krishnan, and Deqing Sun.
\newblock Pyramid adversarial training improves vit performance.
\newblock In \emph{CVPR}, pp.\  13419--13429, 2022.

\bibitem[Houlsby et~al.(2019)Houlsby, Giurgiu, Jastrzebski, Morrone,
  De~Laroussilhe, Gesmundo, Attariyan, and Gelly]{houlsby2019parameter}
Neil Houlsby, Andrei Giurgiu, Stanislaw Jastrzebski, Bruna Morrone, Quentin
  De~Laroussilhe, Andrea Gesmundo, Mona Attariyan, and Sylvain Gelly.
\newblock Parameter-efficient transfer learning for nlp.
\newblock In \emph{ICML}, pp.\  2790--2799. PMLR, 2019.

\bibitem[Huang et~al.(2016)Huang, Sun, Liu, Sedra, and
  Weinberger]{huang2016deep}
Gao Huang, Yu~Sun, Zhuang Liu, Daniel Sedra, and Kilian~Q Weinberger.
\newblock Deep networks with stochastic depth.
\newblock In \emph{ECCV}, pp.\  646--661. Springer, 2016.

\bibitem[Huang et~al.(2020)Huang, Zhang, and Zhang]{huang_self-adaptive_2020}
Lang Huang, Chao Zhang, and Hongyang Zhang.
\newblock Self-{Adaptive} {Training}: beyond {Empirical} {Risk} {Minimization}.
\newblock \emph{arXiv preprint arXiv:2002.10319}, 2020.

\bibitem[Ioffe \& Szegedy(2015)Ioffe and Szegedy]{ioffe2015batch}
Sergey Ioffe and Christian Szegedy.
\newblock Batch normalization: Accelerating deep network training by reducing
  internal covariate shift.
\newblock In \emph{International conference on machine learning}, pp.\
  448--456. PMLR, 2015.

\bibitem[Izmailov et~al.(2018)Izmailov, Podoprikhin, Garipov, Vetrov, and
  Wilson]{izmailov_averaging_2018}
Pavel Izmailov, Dmitrii Podoprikhin, Timur Garipov, Dmitry Vetrov, and
  Andrew~Gordon Wilson.
\newblock Averaging {Weights} {Leads} to {Wider} {Optima} and {Better}
  {Generalization}.
\newblock \emph{Uncertainty in Artificial Intelligence}, 2018.

\bibitem[Kannan et~al.(2018)Kannan, Kurakin, and
  Goodfellow]{kannan_adversarial_2018}
Harini Kannan, Alexey Kurakin, and Ian Goodfellow.
\newblock Adversarial {Logit} {Pairing}.
\newblock \emph{arXiv preprint arXiv:1803.06373}, 2018.

\bibitem[Kireev et~al.(2021)Kireev, Andriushchenko, and
  Flammarion]{kireev_effectiveness_2021}
Klim Kireev, Maksym Andriushchenko, and Nicolas Flammarion.
\newblock On the effectiveness of adversarial training against common
  corruptions.
\newblock \emph{arXiv preprint arXiv:2103.02325}, 2021.
\newblock URL \url{https://arxiv.org/pdf/2103.02325}.

\bibitem[Kolesnikov et~al.(2020)Kolesnikov, Beyer, Zhai, Puigcerver, Yung,
  Gelly, and Houlsby]{kolesnikov2020big}
Alexander Kolesnikov, Lucas Beyer, Xiaohua Zhai, Joan Puigcerver, Jessica Yung,
  Sylvain Gelly, and Neil Houlsby.
\newblock Big transfer (bit): General visual representation learning.
\newblock In \emph{ECCV}, pp.\  491--507. Springer, 2020.

\bibitem[Krizhevsky et~al.(2014)Krizhevsky, Nair, and
  Hinton]{krizhevsky_cifar-10_2014}
Alex Krizhevsky, Vinod Nair, and Geoffrey Hinton.
\newblock The {CIFAR}-10 dataset.
\newblock 2014.
\newblock URL \url{http://www.cs.toronto.edu/kriz/cifar.html}.

\bibitem[Kurakin et~al.(2016{\natexlab{a}})Kurakin, Goodfellow, and
  Bengio]{kurakin2016adversarial}
Alexey Kurakin, Ian Goodfellow, and Samy Bengio.
\newblock Adversarial machine learning at scale.
\newblock \emph{arXiv preprint arXiv:1611.01236}, 2016{\natexlab{a}}.

\bibitem[Kurakin et~al.(2016{\natexlab{b}})Kurakin, Goodfellow, and
  Bengio]{kurakin_adversarial_2016}
Alexey Kurakin, Ian Goodfellow, and Samy Bengio.
\newblock Adversarial examples in the physical world.
\newblock \emph{ICLR workshop}, 2016{\natexlab{b}}.

\bibitem[LeCun et~al.(2010)LeCun, Cortes, and Burges]{lecun2010mnist}
Yann LeCun, Corinna Cortes, and CJ~Burges.
\newblock Mnist handwritten digit database.
\newblock \emph{ATT Labs [Online]. Available:
  http://yann.lecun.com/exdb/mnist}, 2, 2010.

\bibitem[Li et~al.(2016)Li, Wang, Shi, Liu, and Hou]{li2016revisiting}
Yanghao Li, Naiyan Wang, Jianping Shi, Jiaying Liu, and Xiaodi Hou.
\newblock Revisiting batch normalization for practical domain adaptation.
\newblock \emph{arXiv preprint arXiv:1603.04779}, 2016.

\bibitem[Loshchilov \& Hutter(2017)Loshchilov and
  Hutter]{loshchilov2017decoupled}
Ilya Loshchilov and Frank Hutter.
\newblock Decoupled weight decay regularization.
\newblock \emph{arXiv preprint arXiv:1711.05101}, 2017.

\bibitem[Madry et~al.(2018)Madry, Makelov, Schmidt, Tsipras, and
  Vladu]{madry_towards_2017}
Aleksander Madry, Aleksandar Makelov, Ludwig Schmidt, Dimitris Tsipras, and
  Adrian Vladu.
\newblock Towards deep learning models resistant to adversarial attacks.
\newblock \emph{ICLR}, 2018.

\bibitem[Mallya et~al.(2018)Mallya, Davis, and Lazebnik]{mallya2018piggyback}
Arun Mallya, Dillon Davis, and Svetlana Lazebnik.
\newblock Piggyback: Adapting a single network to multiple tasks by learning to
  mask weights.
\newblock In \emph{ECCV}, pp.\  67--82, 2018.

\bibitem[Mancini et~al.(2018)Mancini, Ricci, Caputo, and
  Rota~Bulo]{mancini2018adding}
Massimiliano Mancini, Elisa Ricci, Barbara Caputo, and Samuel Rota~Bulo.
\newblock Adding new tasks to a single network with weight transformations
  using binary masks.
\newblock In \emph{Proceedings of the European Conference on Computer Vision
  (ECCV) Workshops}, pp.\  0--0, 2018.

\bibitem[Maria~Carlucci et~al.(2017)Maria~Carlucci, Porzi, Caputo, Ricci, and
  Rota~Bulo]{maria2017autodial}
Fabio Maria~Carlucci, Lorenzo Porzi, Barbara Caputo, Elisa Ricci, and Samuel
  Rota~Bulo.
\newblock Autodial: Automatic domain alignment layers.
\newblock In \emph{ICCV}, pp.\  5067--5075, 2017.

\bibitem[Netzer et~al.(2011)Netzer, Wang, Coates, Bissacco, Wu, and
  Ng]{netzer2011reading}
Yuval Netzer, Tao Wang, Adam Coates, Alessandro Bissacco, Bo~Wu, and Andrew~Y
  Ng.
\newblock Reading digits in natural images with unsupervised feature learning.
\newblock 2011.

\bibitem[Ovadia et~al.(2019)Ovadia, Fertig, Ren, Nado, Sculley, Nowozin,
  Dillon, Lakshminarayanan, and Snoek]{ovadia2019can}
Yaniv Ovadia, Emily Fertig, Jie Ren, Zachary Nado, David Sculley, Sebastian
  Nowozin, Joshua Dillon, Balaji Lakshminarayanan, and Jasper Snoek.
\newblock Can you trust your model's uncertainty? evaluating predictive
  uncertainty under dataset shift.
\newblock \emph{NeurIPS}, 32, 2019.

\bibitem[Pang et~al.(2020)Pang, Yang, Dong, Xu, Su, and
  Zhu]{pang_boosting_2020}
Tianyu Pang, Xiao Yang, Yinpeng Dong, Kun Xu, Hang Su, and Jun Zhu.
\newblock Boosting {Adversarial} {Training} with {Hypersphere} {Embedding}.
\newblock \emph{NeurIPS}, 2020.

\bibitem[Papernot et~al.(2016)Papernot, McDaniel, Wu, Jha, and
  Swami]{papernot_distillation_2015}
Nicolas Papernot, Patrick McDaniel, Xi~Wu, Somesh Jha, and Ananthram Swami.
\newblock Distillation as a defense to adversarial perturbations against deep
  neural networks.
\newblock \emph{IEEE Symposium on Security and Privacy}, 2016.

\bibitem[Perez et~al.(2018)Perez, Strub, De~Vries, Dumoulin, and
  Courville]{perez2018film}
Ethan Perez, Florian Strub, Harm De~Vries, Vincent Dumoulin, and Aaron
  Courville.
\newblock Film: Visual reasoning with a general conditioning layer.
\newblock In \emph{Proceedings of the AAAI Conference on Artificial
  Intelligence}, volume~32, 2018.

\bibitem[Pfeiffer et~al.(2020)Pfeiffer, Kamath, R{\"u}ckl{\'e}, Cho, and
  Gurevych]{pfeiffer2020adapterfusion}
Jonas Pfeiffer, Aishwarya Kamath, Andreas R{\"u}ckl{\'e}, Kyunghyun Cho, and
  Iryna Gurevych.
\newblock Adapterfusion: Non-destructive task composition for transfer
  learning.
\newblock \emph{arXiv preprint arXiv:2005.00247}, 2020.

\bibitem[Rebuffi et~al.(2017)Rebuffi, Bilen, and Vedaldi]{rebuffi2017learning}
Sylvestre-Alvise Rebuffi, Hakan Bilen, and Andrea Vedaldi.
\newblock Learning multiple visual domains with residual adapters.
\newblock \emph{NeurIPS}, 30, 2017.

\bibitem[Rebuffi et~al.(2021)Rebuffi, Gowal, Calian, Stimberg, Wiles, and
  Mann]{rebuffi2021data}
Sylvestre-Alvise Rebuffi, Sven Gowal, Dan~Andrei Calian, Florian Stimberg,
  Olivia Wiles, and Timothy~A Mann.
\newblock Data augmentation can improve robustness.
\newblock \emph{NeurIPS}, 34:\penalty0 29935--29948, 2021.

\bibitem[Rice et~al.(2020)Rice, Wong, and Kolter]{rice_overfitting_2020}
Leslie Rice, Eric Wong, and J.~Zico Kolter.
\newblock Overfitting in adversarially robust deep learning.
\newblock \emph{ICML}, 2020.

\bibitem[Rosenfeld \& Tsotsos(2018)Rosenfeld and
  Tsotsos]{rosenfeld2018incremental}
Amir Rosenfeld and John~K Tsotsos.
\newblock Incremental learning through deep adaptation.
\newblock \emph{IEEE TPAMI}, 42\penalty0 (3):\penalty0 651--663, 2018.

\bibitem[Russakovsky et~al.(2015)Russakovsky, Deng, Su, Krause, Satheesh, Ma,
  Huang, Karpathy, Khosla, Bernstein, et~al.]{russakovsky2015imagenet}
Olga Russakovsky, Jia Deng, Hao Su, Jonathan Krause, Sanjeev Satheesh, Sean Ma,
  Zhiheng Huang, Andrej Karpathy, Aditya Khosla, Michael Bernstein, et~al.
\newblock Imagenet large scale visual recognition challenge.
\newblock \emph{International journal of computer vision}, 2015.

\bibitem[Stickland \& Murray(2019)Stickland and Murray]{stickland2019bert}
Asa~Cooper Stickland and Iain Murray.
\newblock Bert and pals: Projected attention layers for efficient adaptation in
  multi-task learning.
\newblock In \emph{ICML}, pp.\  5986--5995. PMLR, 2019.

\bibitem[Szegedy et~al.(2014)Szegedy, Zaremba, Sutskever, Bruna, Erhan,
  Goodfellow, and Fergus]{szegedy_intriguing_2013}
Christian Szegedy, Wojciech Zaremba, Ilya Sutskever, Joan Bruna, Dumitru Erhan,
  Ian Goodfellow, and Rob Fergus.
\newblock Intriguing properties of neural networks.
\newblock \emph{ICLR}, 2014.

\bibitem[Tsipras et~al.(2018)Tsipras, Santurkar, Engstrom, Turner, and
  Madry]{tsipras_robustness_2018}
Dimitris Tsipras, Shibani Santurkar, Logan Engstrom, Alexander Turner, and
  Aleksander Madry.
\newblock Robustness may be at odds with accuracy.
\newblock \emph{arXiv preprint arXiv:1805.12152}, 2018.

\bibitem[Uesato et~al.(2019)Uesato, Alayrac, Huang, Stanforth, Fawzi, and
  Kohli]{uesato_are_2019}
Jonathan Uesato, Jean-Baptiste Alayrac, Po-Sen Huang, Robert Stanforth,
  Alhussein Fawzi, and Pushmeet Kohli.
\newblock Are labels required for improving adversarial robustness?
\newblock \emph{NeurIPS}, 2019.

\bibitem[Vaswani et~al.(2017)Vaswani, Shazeer, Parmar, Uszkoreit, Jones, Gomez,
  Kaiser, and Polosukhin]{vaswani2017attention}
Ashish Vaswani, Noam Shazeer, Niki Parmar, Jakob Uszkoreit, Llion Jones,
  Aidan~N Gomez, {\L}ukasz Kaiser, and Illia Polosukhin.
\newblock Attention is all you need.
\newblock \emph{NeurIPS}, 30, 2017.

\bibitem[Walter et~al.(2022)Walter, Stutz, and Schiele]{walter2022fragile}
Nils~Philipp Walter, David Stutz, and Bernt Schiele.
\newblock On fragile features and batch normalization in adversarial training.
\newblock \emph{arXiv preprint arXiv:2204.12393}, 2022.

\bibitem[Wang et~al.(2019)Wang, Ge, Lipton, and Xing]{wang2019learning}
Haohan Wang, Songwei Ge, Zachary Lipton, and Eric~P Xing.
\newblock Learning robust global representations by penalizing local predictive
  power.
\newblock In \emph{Advances in Neural Information Processing Systems}, pp.\
  10506--10518, 2019.

\bibitem[Wang et~al.(2022)Wang, Zhang, Zheng, Shi, Li, and
  Wang]{wang2022removing}
Haotao Wang, Aston Zhang, Shuai Zheng, Xingjian Shi, Mu~Li, and Zhangyang Wang.
\newblock Removing batch normalization boosts adversarial training.
\newblock In \emph{International Conference on Machine Learning}, pp.\
  23433--23445. PMLR, 2022.

\bibitem[Wang et~al.(2020)Wang, Tang, Duan, Wei, Huang, Cao, Jiang, Zhou,
  et~al.]{wang2020k}
Ruize Wang, Duyu Tang, Nan Duan, Zhongyu Wei, Xuanjing Huang, Guihong Cao,
  Daxin Jiang, Ming Zhou, et~al.
\newblock K-adapter: Infusing knowledge into pre-trained models with adapters.
\newblock \emph{arXiv preprint arXiv:2002.01808}, 2020.

\bibitem[Wightman(2019)]{rw2019timm}
Ross Wightman.
\newblock Pytorch image models.
\newblock \url{https://github.com/rwightman/pytorch-image-models}, 2019.

\bibitem[Wong et~al.(2020)Wong, Rice, and Kolter]{wong2020fast}
Eric Wong, Leslie Rice, and J~Zico Kolter.
\newblock Fast is better than free: Revisiting adversarial training.
\newblock \emph{arXiv preprint arXiv:2001.03994}, 2020.

\bibitem[Wortsman et~al.(2022)Wortsman, Ilharco, Gadre, Roelofs, Gontijo-Lopes,
  Morcos, Namkoong, Farhadi, Carmon, Kornblith, et~al.]{wortsman2022model}
Mitchell Wortsman, Gabriel Ilharco, Samir~Ya Gadre, Rebecca Roelofs, Raphael
  Gontijo-Lopes, Ari~S Morcos, Hongseok Namkoong, Ali Farhadi, Yair Carmon,
  Simon Kornblith, et~al.
\newblock Model soups: averaging weights of multiple fine-tuned models improves
  accuracy without increasing inference time.
\newblock In \emph{ICML}, pp.\  23965--23998. PMLR, 2022.

\bibitem[Wu et~al.(2020)Wu, Xia, and Wang]{wu2020adversarial}
Dongxian Wu, Shu-tao Xia, and Yisen Wang.
\newblock Adversarial weight perturbation helps robust generalization.
\newblock \emph{NeurIPS}, 2020.

\bibitem[Wu \& He(2018)Wu and He]{wu2018group}
Yuxin Wu and Kaiming He.
\newblock Group normalization.
\newblock In \emph{ECCV}, pp.\  3--19, 2018.

\bibitem[Xiao et~al.(2010)Xiao, Hays, Ehinger, Oliva, and
  Torralba]{xiao2010sun}
Jianxiong Xiao, James Hays, Krista~A Ehinger, Aude Oliva, and Antonio Torralba.
\newblock Sun database: Large-scale scene recognition from abbey to zoo.
\newblock In \emph{2010 IEEE computer society conference on computer vision and
  pattern recognition}, pp.\  3485--3492. IEEE, 2010.

\bibitem[Xie \& Yuille(2019)Xie and Yuille]{xie2019intriguing}
Cihang Xie and Alan Yuille.
\newblock Intriguing properties of adversarial training at scale.
\newblock \emph{arXiv preprint arXiv:1906.03787}, 2019.

\bibitem[Xie et~al.(2019{\natexlab{a}})Xie, Tan, Gong, Wang, Yuille, and
  Le]{xie_adversarial_2019}
Cihang Xie, Mingxing Tan, Boqing Gong, Jiang Wang, Alan Yuille, and Quoc~V Le.
\newblock Adversarial {Examples} {Improve} {Image} {Recognition}.
\newblock \emph{arXiv preprint arXiv:1911.09665}, 2019{\natexlab{a}}.
\newblock URL \url{https://arxiv.org/abs/1911.09665}.

\bibitem[Xie et~al.(2019{\natexlab{b}})Xie, Wu, van~der Maaten, Yuille, and
  He]{xie_feature_2018}
Cihang Xie, Yuxin Wu, Laurens van~der Maaten, Alan Yuille, and Kaiming He.
\newblock Feature denoising for improving adversarial robustness.
\newblock \emph{CVPR}, 2019{\natexlab{b}}.

\bibitem[Yun et~al.(2019)Yun, Han, Oh, Chun, Choe, and Yoo]{yun2019cutmix}
Sangdoo Yun, Dongyoon Han, Seong~Joon Oh, Sanghyuk Chun, Junsuk Choe, and
  Youngjoon Yoo.
\newblock Cutmix: Regularization strategy to train strong classifiers with
  localizable features.
\newblock \emph{ICCV}, 2019.

\bibitem[Zhang et~al.(2019)Zhang, Yu, Jiao, Xing, Ghaoui, and
  Jordan]{zhang_theoretically_2019}
Hongyang Zhang, Yaodong Yu, Jiantao Jiao, Eric~P. Xing, Laurent~El Ghaoui, and
  Michael~I. Jordan.
\newblock Theoretically {Principled} {Trade}-off between {Robustness} and
  {Accuracy}.
\newblock \emph{ICML}, 2019.

\bibitem[Zhang et~al.(2018)Zhang, Cisse, Dauphin, and
  Lopez-Paz]{zhang2017mixup}
Hongyi Zhang, Moustapha Cisse, Yann~N Dauphin, and David Lopez-Paz.
\newblock mixup: Beyond empirical risk minimization.
\newblock \emph{ICLR}, 2018.

\bibitem[Zhang et~al.(2021)Zhang, Zhang, Zhao, Chen, and
  Pfister]{zhang2021aggregating}
Zizhao Zhang, Han Zhang, Long Zhao, Ting Chen, and Tomas Pfister.
\newblock Aggregating nested transformers.
\newblock \emph{arXiv preprint arXiv:2105.12723}, 2021.

\end{thebibliography}
\bibliographystyle{iclr2023_conference}

\clearpage

\appendix
\section{More experimental details}

\paragraph{Training details.} In this manuscript we train \vitb models using the training pipeline proposed in \cite{he2022masked}. The model is optimized for 300 epochs using the AdamW optimizer \citep{loshchilov2017decoupled} with momenta $\beta_1=0.9$, $\beta_2=0.95$, with a weight decay of 0.3 and a cosine learning rate decay with base learning rate 1e-4 and linear ramp-up of 20 epochs. The batch size is set to 4096 and we scale the learning rates using the linear scaling
rule of \citet{goyal2017accurate}. We optimize the standard cross-entropy loss and we use a label smoothing of 0.1. We apply stochastic depth \citep{huang2016deep} with base value 0.1 and with a dropping probability linearly increasing with depth. Regarding data augmentation, we use random crops resized to 224 $\times$ 224 images, mixup \citep{zhang2017mixup}, CutMix \citep{yun2019cutmix} and RandAugment \citep{cubuk2019randaugment} with two layers, magnitude 9 and a random probability of 0.5. We note that our implementation of RandAugment is based on the version found in the \emph{timm} library~\citep{rw2019timm}. We also use exponential moving average with momentum 0.9999.
For \resnet-50 we keep the same training scheme used for \vitb, and the standard architecture, except for combining GroupNorm with Weight Standardization in all convolutional layers following \citet{kolesnikov2020big}. For the DualParams BatchNorm version we fix the \emph{robust} branch to always use the accumulated statistics rather then the batch ones.

\paragraph{Training on smaller datasets.} When training from scratch on smaller datasets in Section \ref{subsec:small_datasets}, we optimize the smaller \vits with a batch size of 1024 and a base learning rate of 2e-4. For datasets with small image resolution such as \cifar, we do not rescale the images to 224 $\times$ 224 but we use a patch size of 4 and a stride of 2 to get enough vision tokens.

\paragraph{Attack details.} For \pgd{2} and \pgd{5} we use a gradient descent update with a fixed step size of 2.5/255 and 1/255 respectively. For \pgd{40} we change the optimizer to Adam with step-size 0.1 decayed by 10 $\times$ at steps 20 and 30. Regarding one step attacks, we use a step size of 6/255 and initialization radius of 8/255 for N-FGSM and a step size of 5/255 for Fast-AT.

\section{Visualizing filters}

\paragraph{Visualization procedure.} We visualize the embedding layer by first standardizing the weights to have zero mean and unit variance. We then extract the first 28 principal components. Finally we reshape them to 16 $\times$ 16 $\times$ 3 images and rescale them to have their values between 0 and 255 such as to display these components as RGB images.

\begin{figure}[ht]
\centering
\subfigure[\pgd{2} but no adapters \label{fig:patchifier_shared}]{\includegraphics[width=0.3\linewidth,trim=3.5cm 2cm 3.5cm 2cm,clip]{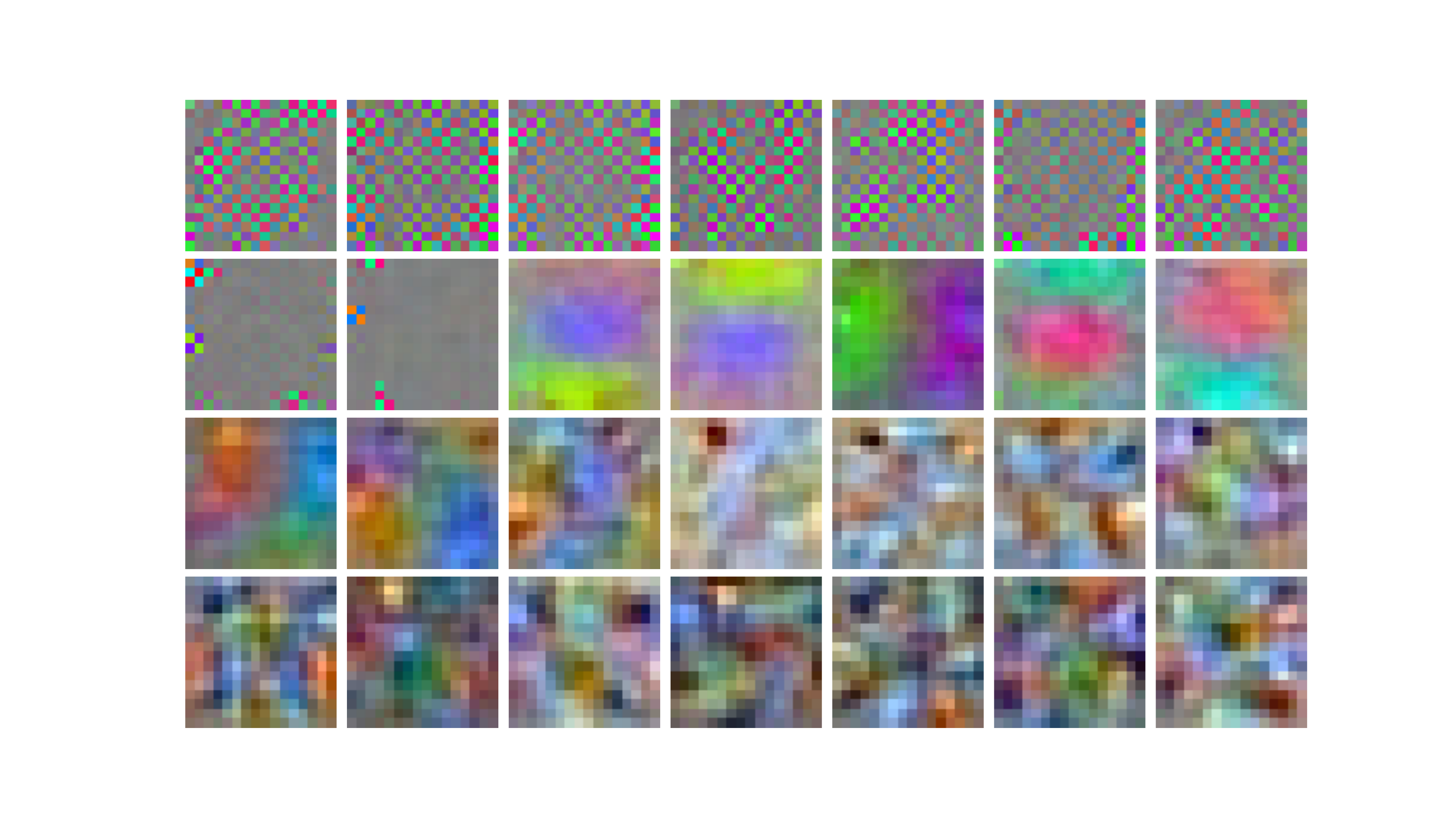}}
\hspace*{1\columnsep}
\subfigure[Fast-AT]{\includegraphics[width=0.3\linewidth,trim=3.5cm 2cm 3.5cm 2cm,clip]{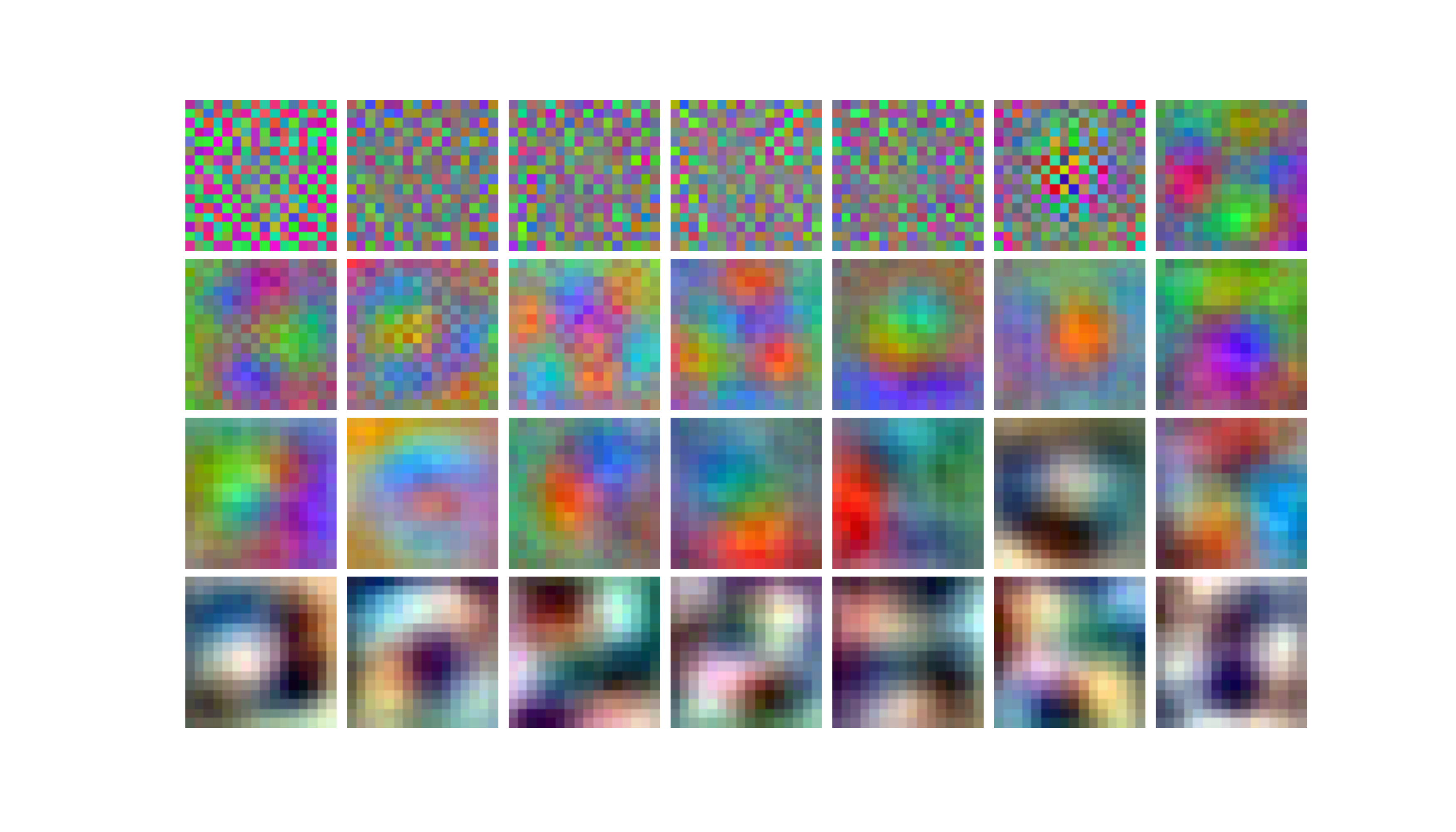}}
\hspace*{1\columnsep}
\subfigure[N-FGSM]{\includegraphics[width=0.3\linewidth,trim=3.5cm 2cm 3.5cm 2cm,clip]{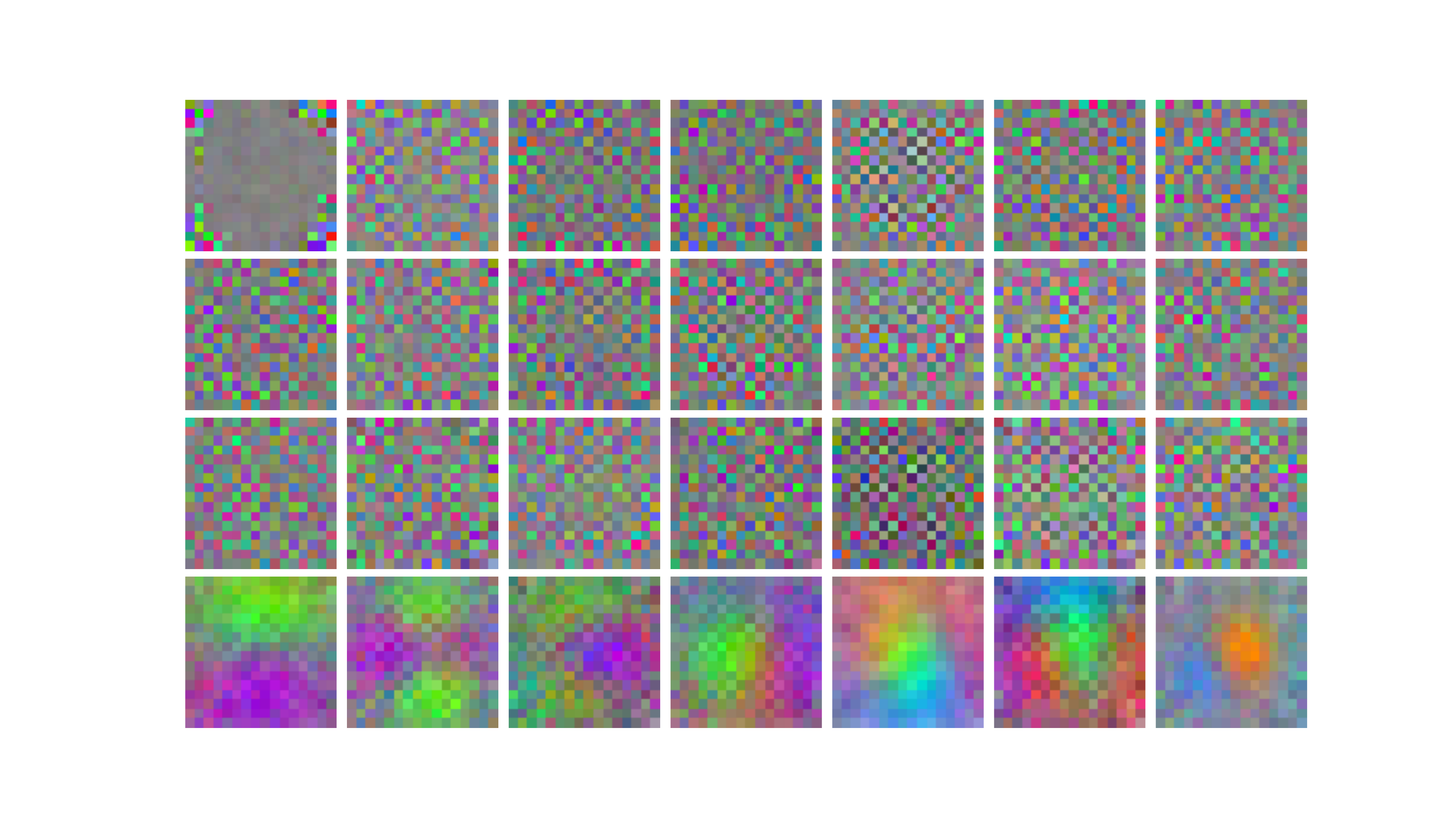}}
\caption{First 28 principal components of the embedding filters of \vitb. In panel (a), a fully shared model (no adapters) is co-trained on clean samples and adversarial samples coming \pgd{2}. In panels (b) and (c), models with classification token adapter are adversarially co-trained with adversarial samples coming from 1 step attacks: Fast-AT and N-FGSM. In these three cases, we observe \emph{catastrophic robust overfitting} with no robustness at the end of training. We observe that there are degenerate filters among the first principal components. \label{fig:filters_1step}}
\end{figure}

\begin{figure}[h]
\centering
\subfigure[ $\alpha=0$ (pure adversarial)]{\includegraphics[width=0.3\linewidth,trim=3.5cm 2cm 3.5cm 2cm,clip]{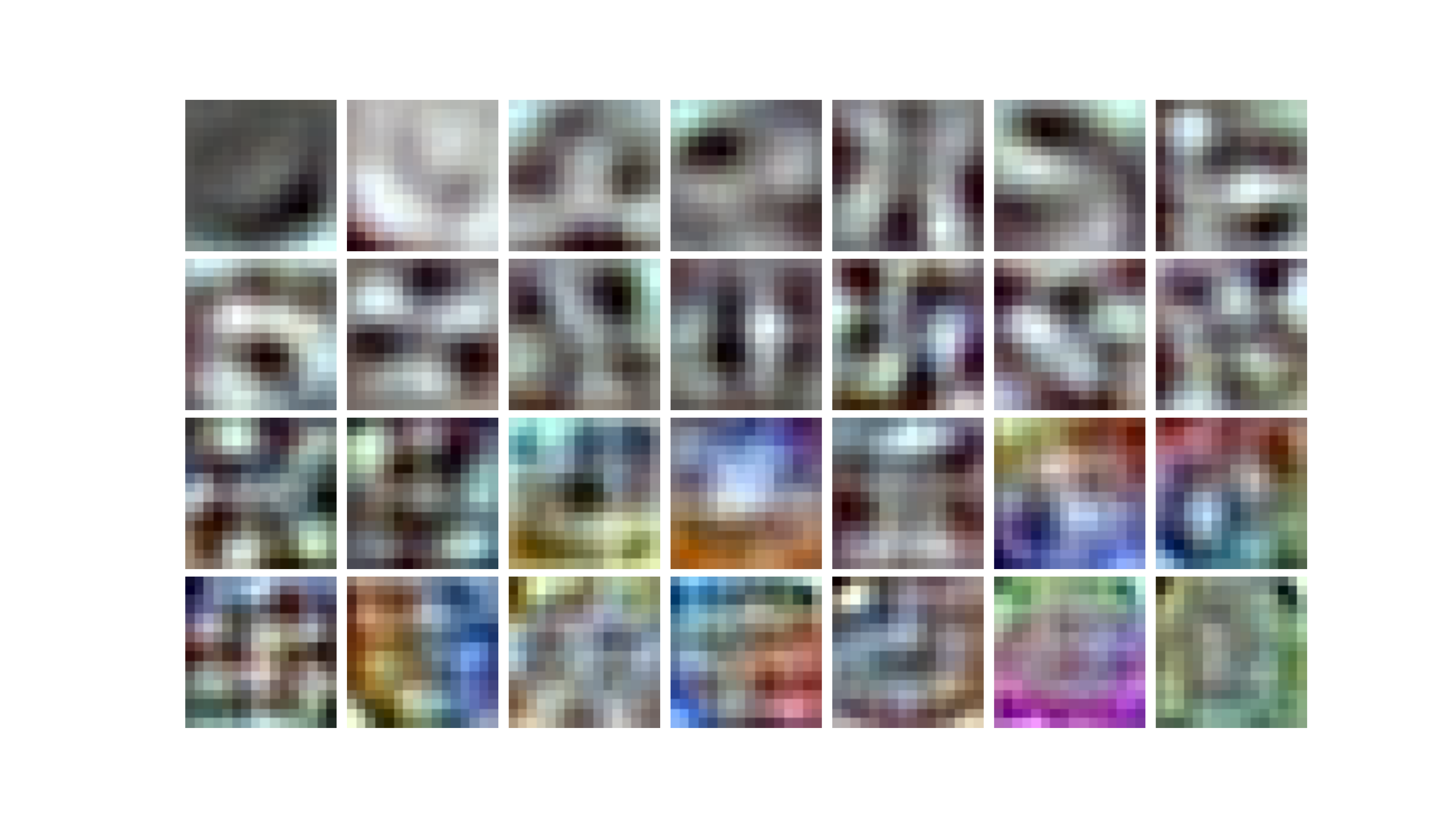}}
\hspace*{1\columnsep}
\subfigure[ $\alpha=0.2$]{\includegraphics[width=0.3\linewidth,trim=3.5cm 2cm 3.5cm 2cm,clip]{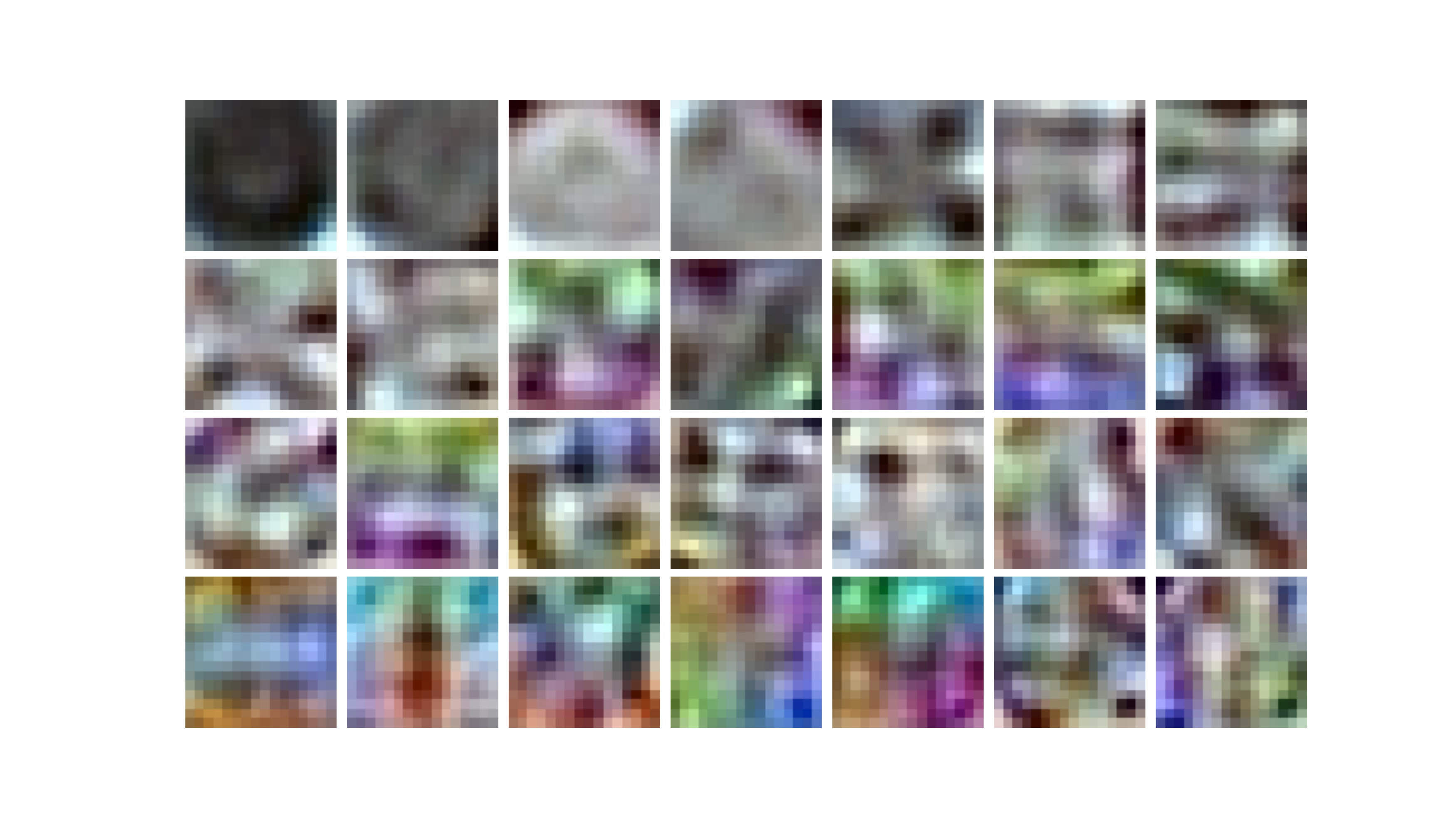}}
\hspace*{1\columnsep}
\subfigure[ $\alpha=0.4$]{\includegraphics[width=0.3\linewidth,trim=3.5cm 2cm 3.5cm 2cm,clip]{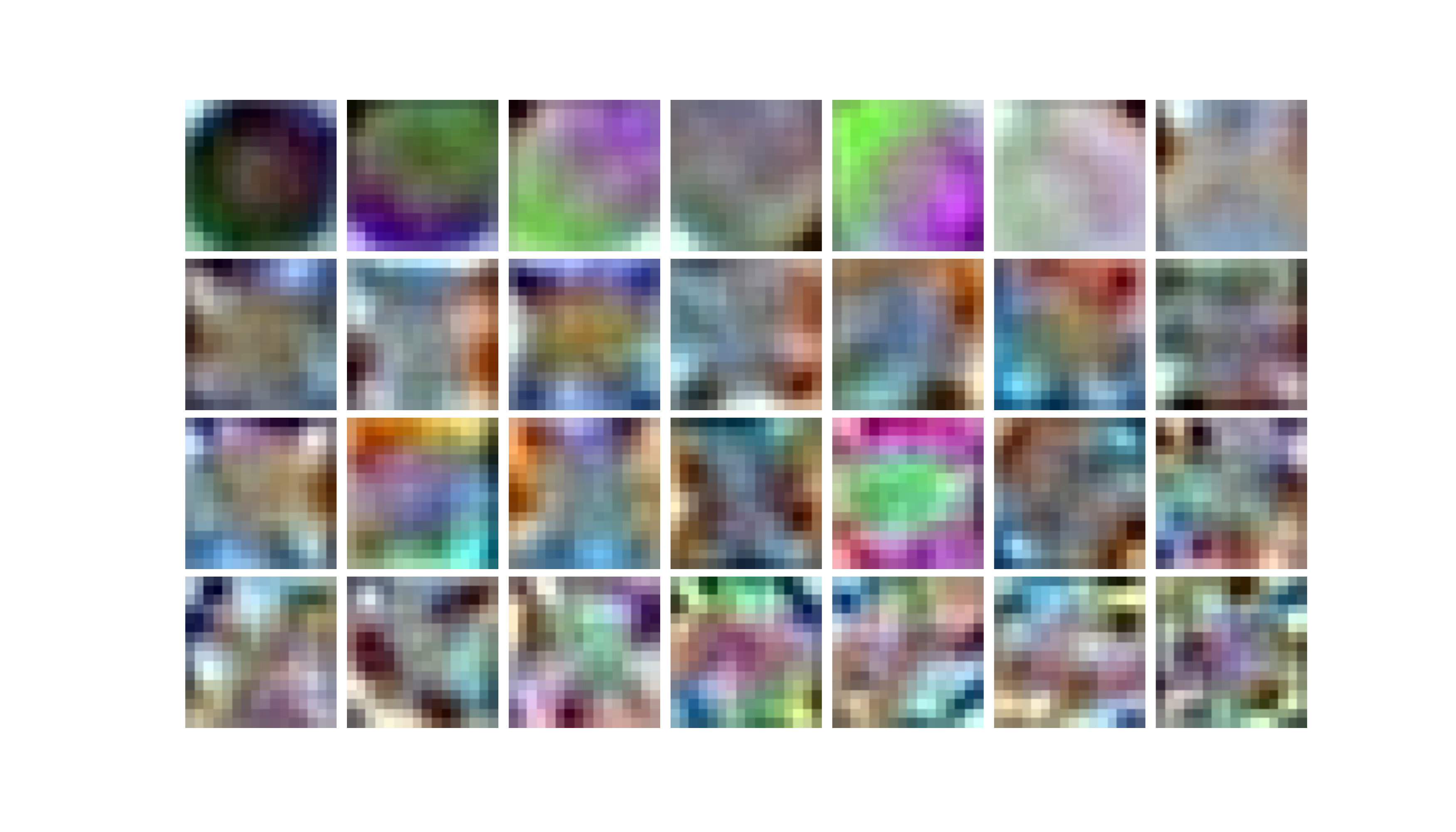}}
\subfigure[ $\alpha=0.6$]{\includegraphics[width=0.3\linewidth,trim=3.5cm 2cm 3.5cm 2cm,clip]{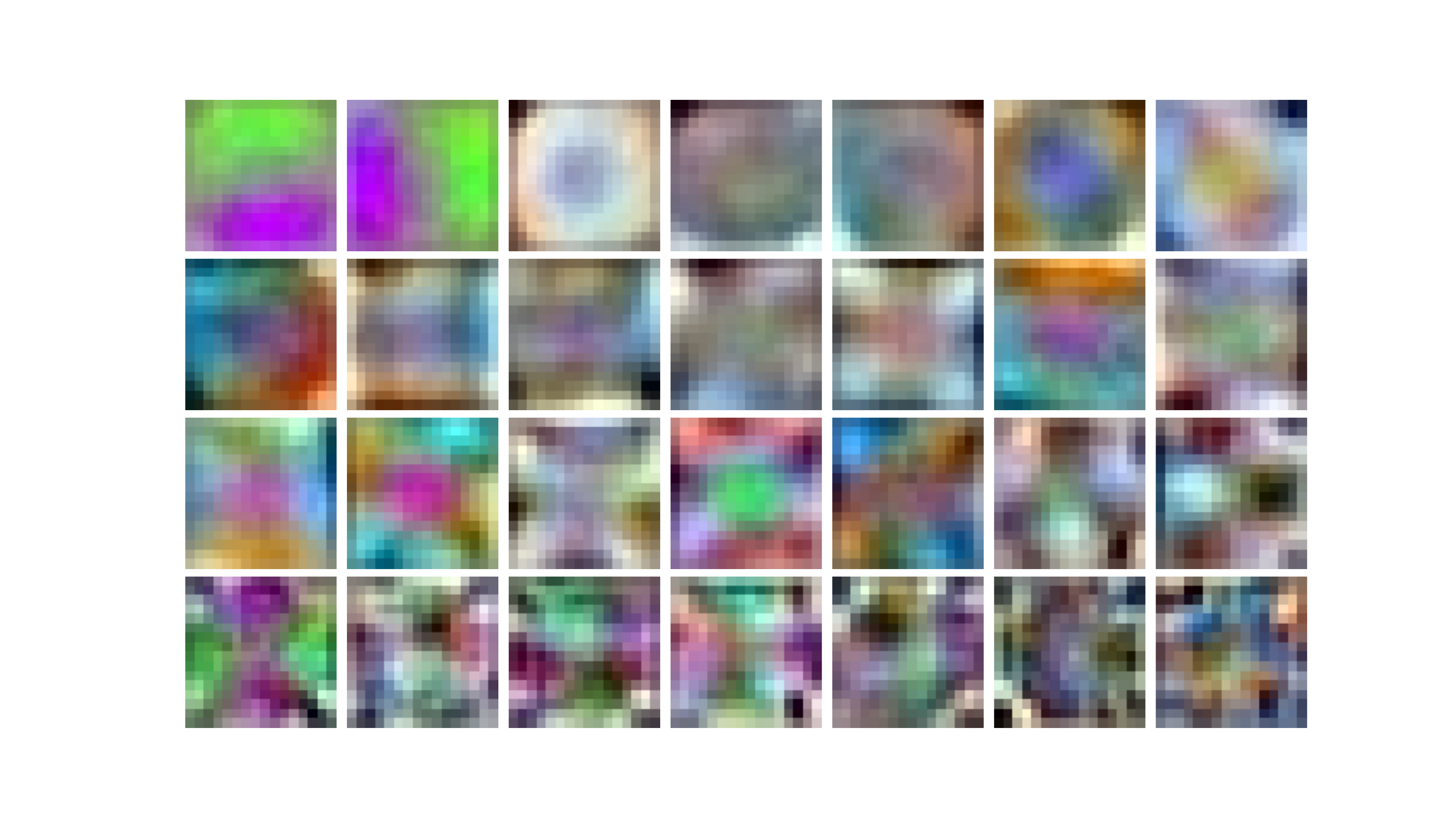}}
\hspace*{1\columnsep}
\subfigure[ $\alpha=0.8$]{\includegraphics[width=0.3\linewidth,trim=3.5cm 2cm 3.5cm 2cm,clip]{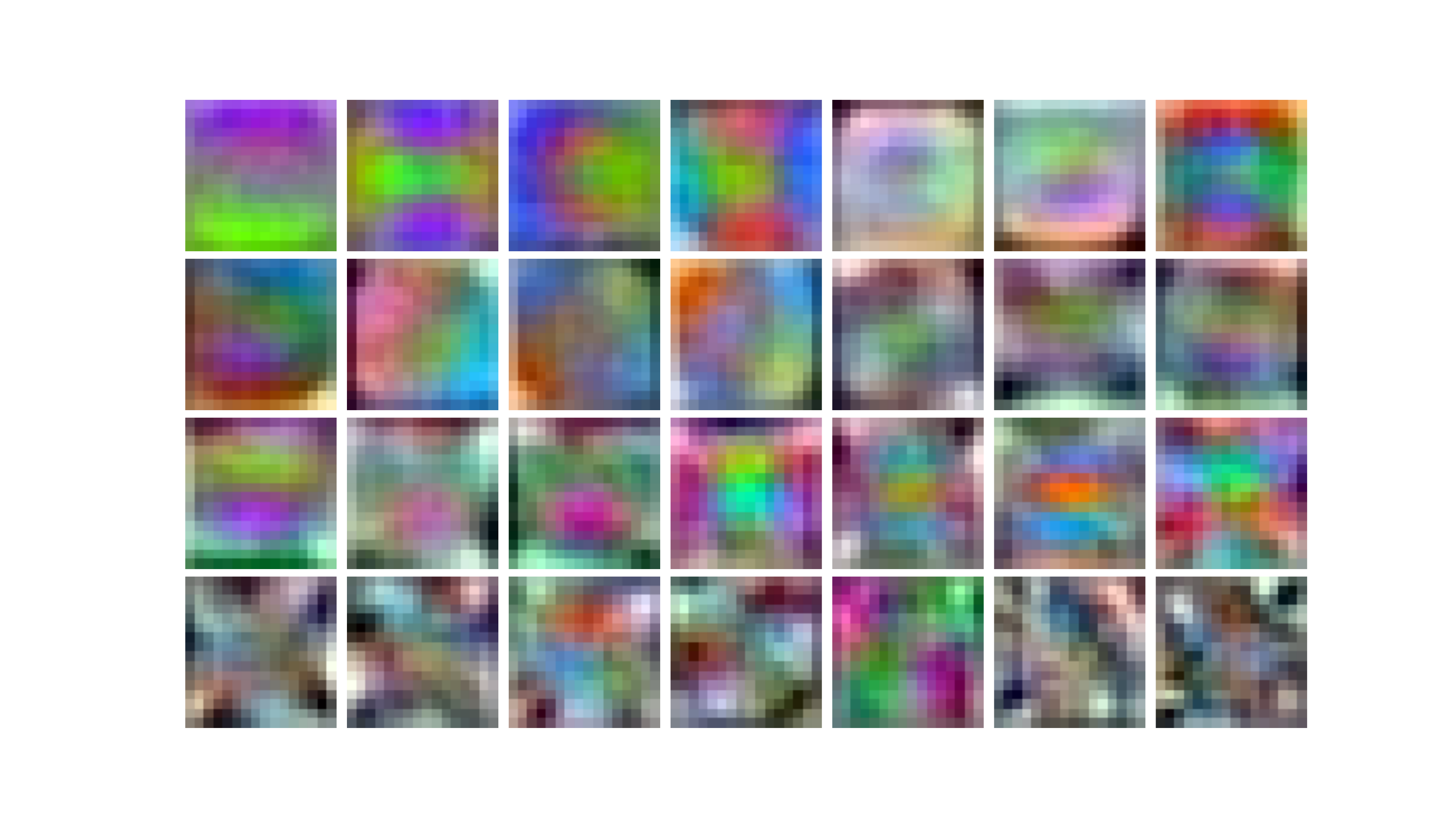}}
\hspace*{1\columnsep}
\subfigure[ $\alpha=1$ (pure nominal)]{\includegraphics[width=0.3\linewidth,trim=3.5cm 2cm 3.5cm 2cm,clip]{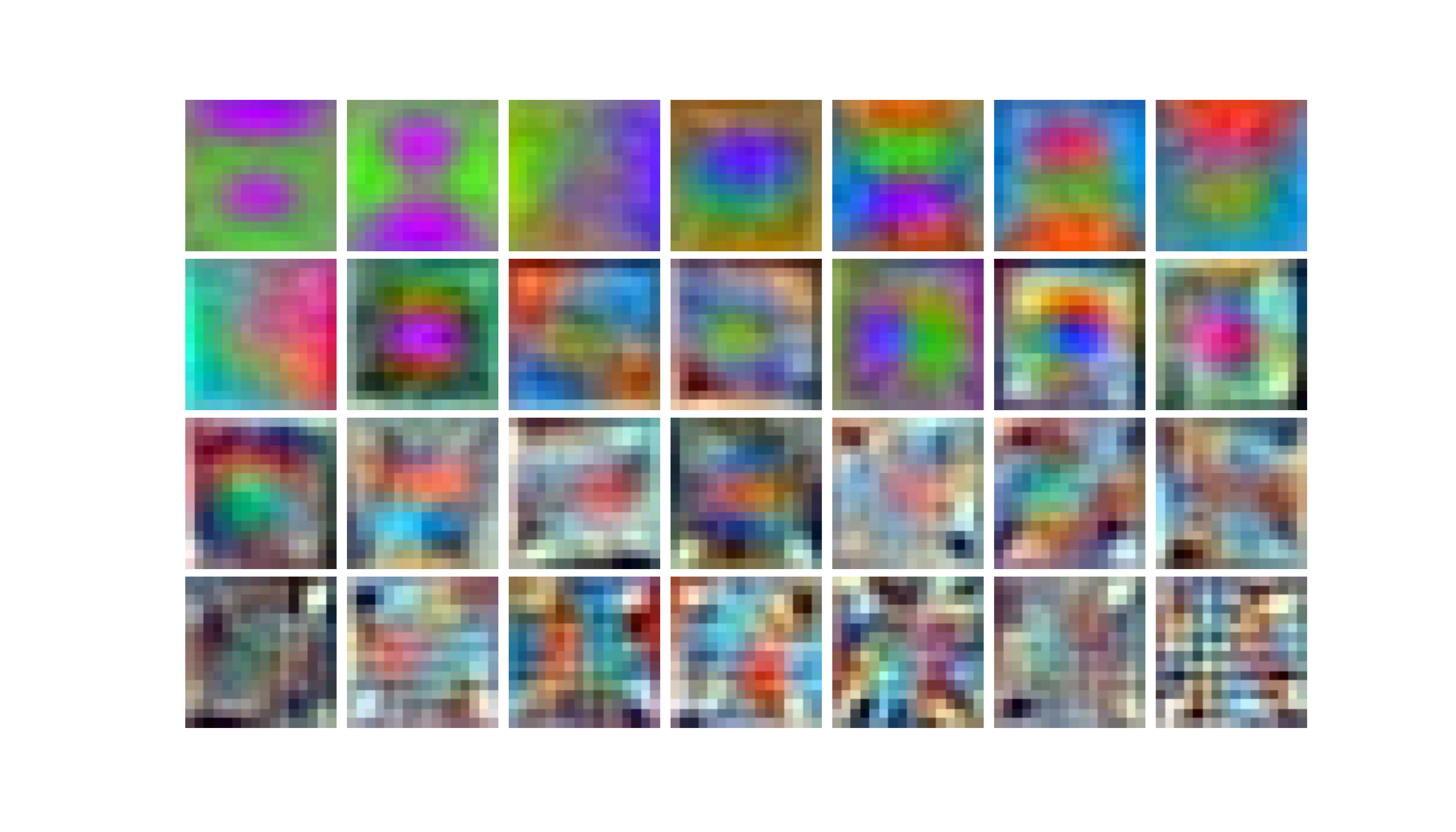}}
\caption{First 28 principal components of the embedding filters of \vitb with classification token adapter trained with various weightings $\alpha$ of the co-training loss Eq. \ref{eq:co_training_loss}. We recall that $\alpha=1$ corresponds to pure nominal training and $\alpha=0$ to adversarial training. Filters learned with $\alpha=0$ focus more on intensity variations (black and white patterns) whereas filters learned with $\alpha=0$ are more color based. Filters with intermediate values of $\alpha$ exhibit circle and star patterns which are not present for $\alpha=0$ or $\alpha=1$. \label{fig:filters}}
\end{figure}

\section{Transfer learning} \label{sec:transfer_learning}

\paragraph{Training details.} For completeness we evaluate the transfer learning performance of the \vitb pre-trained on \imagenet by co-training on clean and adversarial samples. We choose the model trained with classification token adapter and co-training coefficient $\alpha=0.4$, which we fine-tune nominally on \cifar, \cifarh, \textsc{SUN-397}, \textsc{RESISC-45} and \textsc{DMLab} using SGD with momentum 0.9, a batch size of 512, gradient clipping at global norm 1 and no weight decay. We optimize the standard cross-entropy loss and we use a label smoothing of 0.1.  For simplicity, we use the same training schedule for all the datasets: a total of 10k training steps and a base learning rate of 0.01 attained after a linear ramp-up of 500 steps followed by a cosine decay. Regarding data pre-processing, we simply rescale the images to 224 $\times$ 224 resolution without preserving aspect ratio and we apply random horizontal flipping as data augmentation. Finally, we use exponential moving average with momentum 0.999.

\paragraph{Fine-tuning results.} As the network was pre-trained with classification token adapter, we have several possibilities for initializing the classification token before fine-tuning: adversarial token, clean token and \emph{model soups} interpolating between these two tokens. For comparison, we also fine-tune two \vitb pre-trained on \imagenet with nominal and adversarial training respectively. We report the results in Table \ref{tab:finetuning_normal} where we evaluate several fine-tuning strategies: fine-tuning {\it (i)} the classifier head, {\it (ii)} the classifier head and the classification token and {\it (iii)} all the weights. First, we observe that fine-tuning both the classification token and the classifier head brings only a small improvement (from 79.27\% to 80.70\% for the best average accuracy) over fine-tuning the classifier head alone. Fine-tuning all the weights is the best strategy as it reaches 88.40\% average accuracy. Second, we observe that initializing the classification token with the adversarial token performs consistently better than with the clean token when fine-tuning all the weights. Finally, co-training as pre-training is significantly better than nominal and adversarial pre-training as fine-tuning from a co-trained model reaches 88.40\% average accuracy, a +1.05\% improvement over nominal and adversarial pre-training.

\begin{table}[h]
 \caption{\textbf{Co-training as pre-training.} We compare the transfer learning performance of a model pre-trained using co-training to models pre-trained with nominal and adversarial training. We evaluate various fine-tuning strategies on several datasets (headers in green) and we report the average over datasets in the last rows (orange header). We also assess several initializations for the classification token before fine-tuning: adversarial token, clean token and \emph{model soups} between these two tokens with various weightings $\beta$. All models are pre-trained on \imagenet and use the same \vitb architecture during fine-tuning.  \label{tab:finetuning_normal}}%
\begin{center}
\resizebox{1\textwidth}{!}{
\begin{tabular}{l|cc|cccccc}
    \hline
    \cellcolor{header} \textsc{Setup}   & \multicolumn{2}{c|}{\cellcolor{header} \textsc{Baselines}} & \multicolumn{5}{c}{\cellcolor{header} \textsc{From co-trained net}}  \Tstrut \\
    \cellcolor{header}   & \cellcolor{header} Nominal & \cellcolor{header} Adversarial & \cellcolor{header} Robust mode & $\beta=0.25$ & $\beta=0.5$ & $\beta=0.75$ &\cellcolor{header} Clean mode \Tstrut \\
    \hline
    \hline
    \multicolumn{8}{l}{\cellcolor{subheader} \cifar} \TBstrut \\
    \hline
    Fine-tune head  &  96.07\% & 90.95\% & 90.28\% & 91.17\% & 93.61\% & 96.50\% & \textbf{97.15\%} \Tstrut \\
    Fine-tune head + cls token &  96.62\% & 92.76\% & 97.73\% & 97.70\% & 97.77\% & 97.82\% & \textbf{97.84\%} \\
    Fine-tune all &  98.68\% & 98.96\% & \textbf{99.09\%} & 99.03\% & 99.01\% & 99.05\% & 99.03\% \Bstrut\\
    \hline
    \hline
    \multicolumn{8}{l}{\cellcolor{subheader} \cifarh} \TBstrut \\
    \hline
    Fine-tune head  &  83.30\% & 73.80\% & 71.94\% & 73.52\% & 77.78\% & 83.99\% & \textbf{85.47\%} \Tstrut \\
    Fine-tune head + cls token &  84.59\% & 76.79\% & 87.26\% & 87.49\% & \textbf{87.55\%} & 87.45\% & 87.43\% \\
    Fine-tune all &  91.18\% & 91.74\% & 92.37\% & 92.23\% & 92.32\% & \textbf{92.41\%} & 92.29\% \Bstrut\\
    \hline
    \hline
    \multicolumn{8}{l}{\cellcolor{subheader} \textsc{SUN-397}} \TBstrut \\
    \hline
    Fine-tune head  &  72.70\% & 65.62\% & 65.93\% & 67.02\% & 70.19\% & 73.00\% & \textbf{73.47\%} \Tstrut \\
    Fine-tune head + cls token &  73.05\% & 67.21\% & 73.99\% & 74.14\% & \textbf{74.19\%} & 74.12\% & 74.15\% \\
    Fine-tune all &  76.48\% & 75.66\% & \textbf{77.87\%} & 77.75\% & 77.74\% & 77.67\% & 77.72\% \Bstrut\\
    \hline
    \hline
    \multicolumn{8}{l}{\cellcolor{subheader} \textsc{RESISC-45}} \TBstrut \\
    \hline
    Fine-tune head  &  \textbf{91.69\%} & 86.70\% & 86.54\% & 87.37\% & 89.64\% & 90.58\% & 91.12\% \Tstrut \\
    Fine-tune head + cls token &  \textbf{91.95\%} & 87.52\% & 91.04\% & 91.07\% & 91.04\% & 91.49\% & 91.23\% \\
    Fine-tune all &  96.78\% & 96.14\% & \textbf{97.07\%} & 96.72\% & 96.88\% & \textbf{97.07\%} & 96.80\% \Bstrut\\
    \hline
    \hline
    \multicolumn{8}{l}{\cellcolor{subheader} \textsc{DMLab}} \TBstrut \\
    \hline
    Fine-tune head  &  50.02\% & \textbf{50.11\%} & 48.58\% & 48.60\% & 49.08\% & 49.07\% & 49.16\% \Tstrut \\
    Fine-tune head + cls token &  50.91\% & 51.53\% & 50.81\% & 51.79\% & 52.47\% & \textbf{52.64\%} & 52.41\% \\
    Fine-tune all &  73.65\% & 73.93\% & 75.61\% & 75.66\% & \textbf{75.74\%} & 75.35\% & 75.58\% \Bstrut\\
    \hline
    \hline
    \multicolumn{8}{l}{\cellcolor{subheader2} \textsc{AVERAGE}} \TBstrut \\
    \hline
    Fine-tune head  &  78.76\% & 73.44\% & 72.65\% & 73.54\% & 76.06\% & 78.63\% & \textbf{79.27\%} \Tstrut \\
    Fine-tune head + cls token &  79.42\% & 75.16\% & 80.17\% & 80.44\% & 80.60\% & \textbf{80.70\%} & 80.61\%  \\
    Fine-tune all &  87.35\% & 87.29\% & \textbf{88.40\%} & 88.28\% & 88.34\% & 88.31\% & 88.28\% \Bstrut\\
    \hline
    \end{tabular}
}
\end{center}
\end{table}

\section{Accuracy landscape}

\begin{figure}[t]
\centering
\subfigure[Clean accuracy]{\includegraphics[width=0.32\linewidth,trim=0.5cm 1cm 0.5cm 1cm,clip]{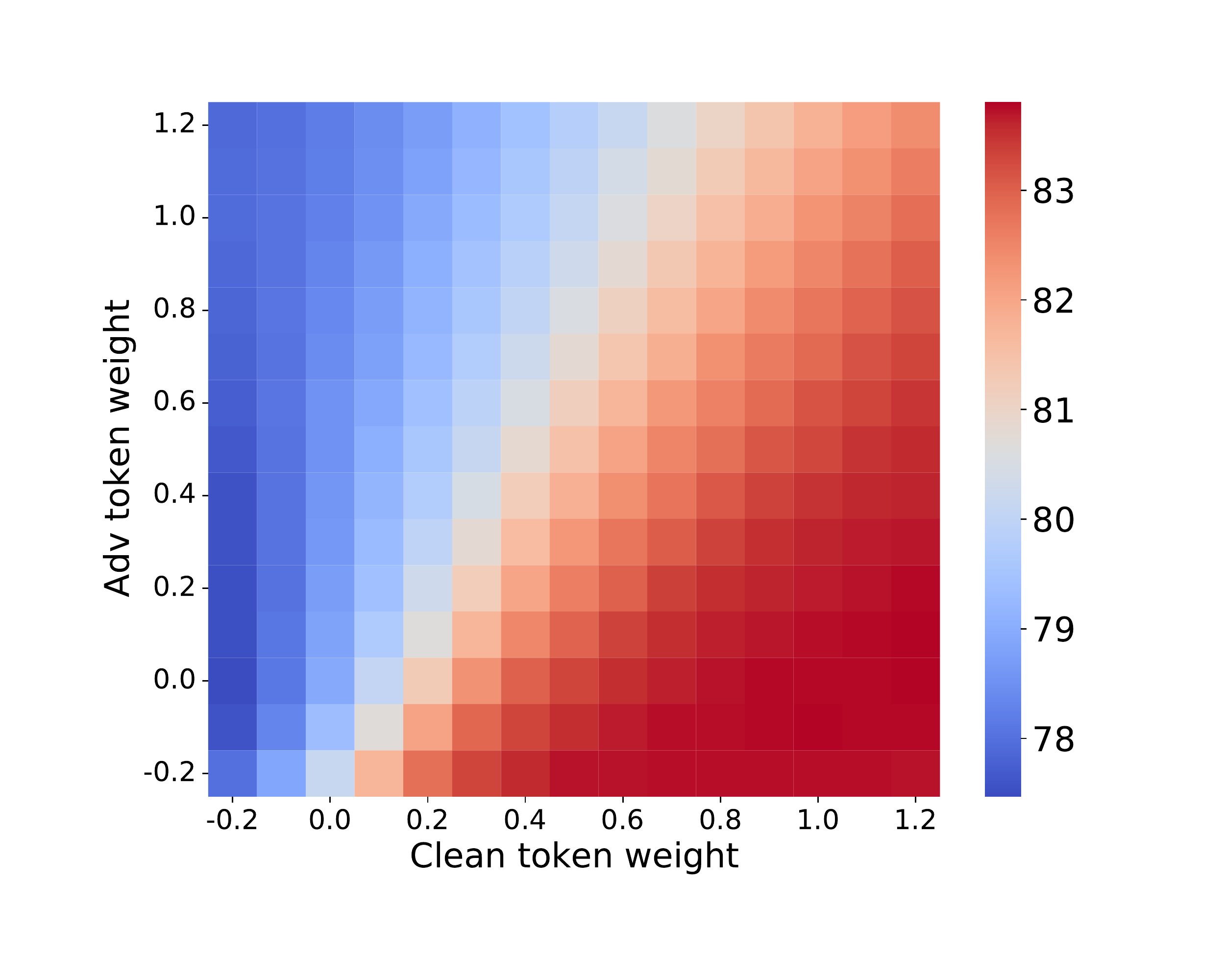}}
\subfigure[Robust accuracy (\pgd{2})]{\includegraphics[width=0.32\linewidth,trim=0.5cm 1cm 0.5cm 1cm,clip]{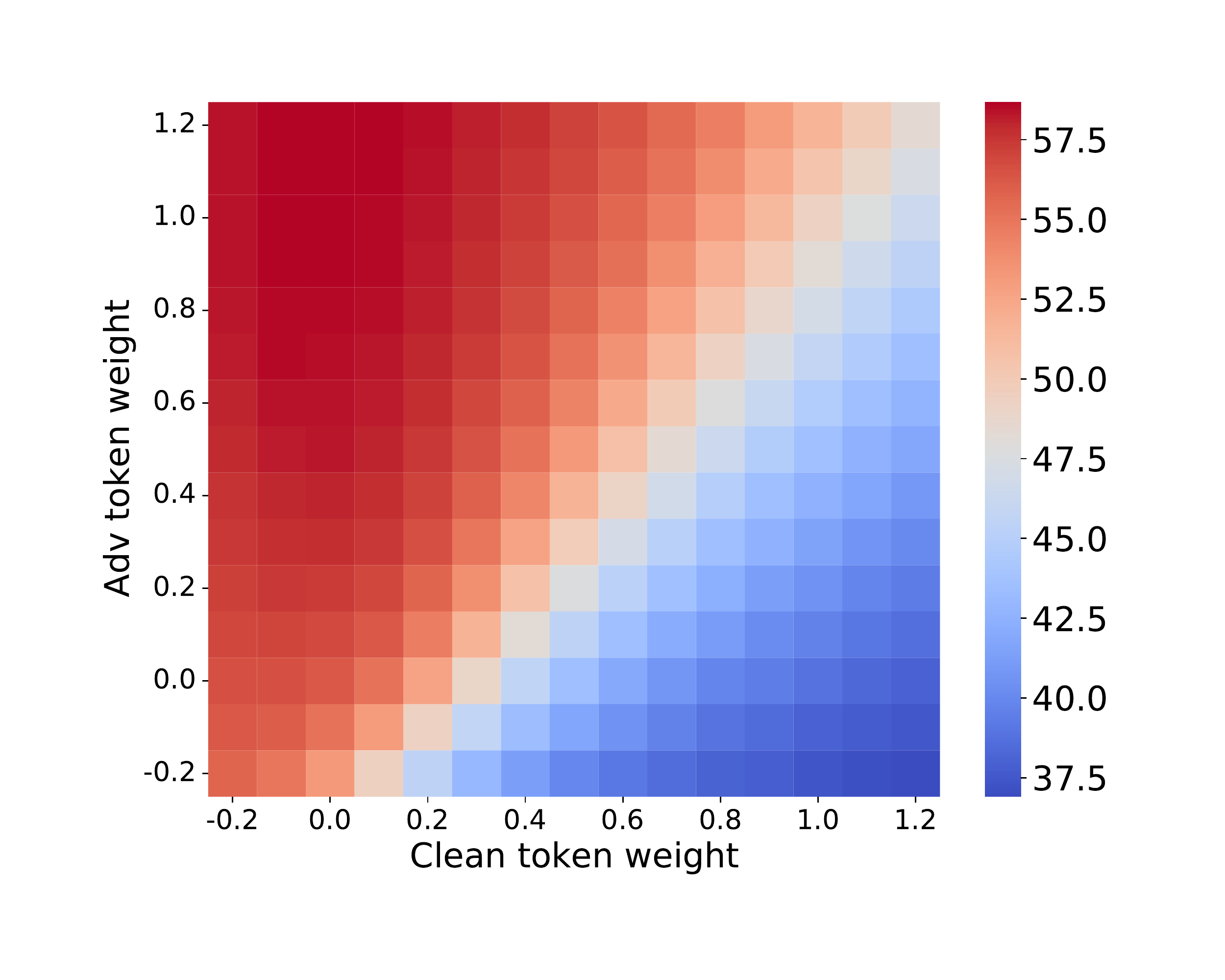}}
\subfigure[Arithmetic mean]{\includegraphics[width=0.29\linewidth,trim=0.5cm 1cm 0.5cm 1cm,clip]{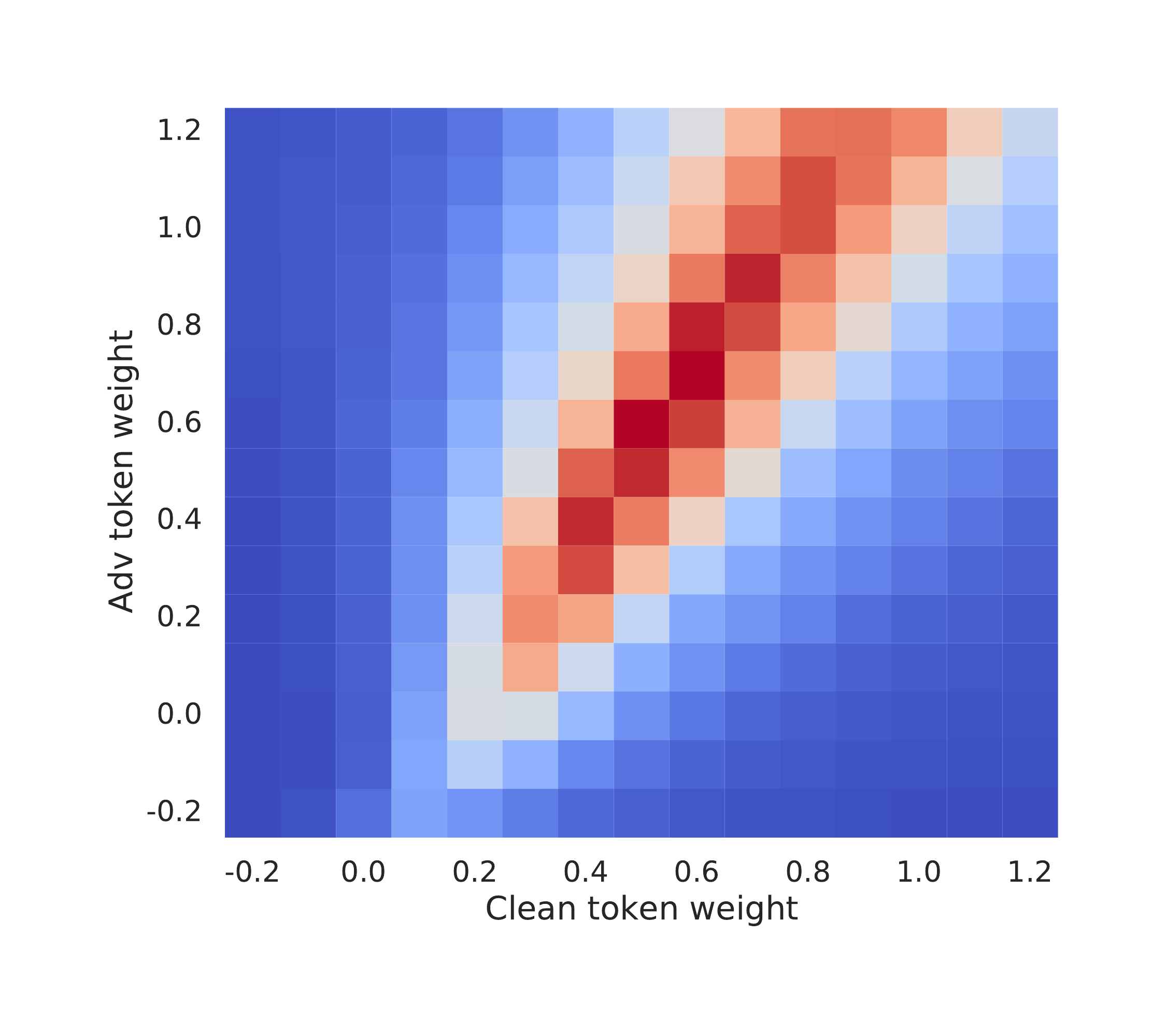}}
\caption{{\bf Linear combination of tokens.} We report the clean accuracy (panel (a)) and robust accuracy against \pgd{2} (panel (b)) on \imagenet  for various linear combinations of the clean and adversarial tokens. \emph{Model soups}, which are linear interpolation (and extrapolation) between these two tokens, are on the diagonal from top left to bottom right. Panel (c) shows the arithmetic mean between the normalized (with min/max rescaling) clean and robust accuracies (red means higher mean accuracy). \label{fig:2d_landscape}}
\end{figure}

In our case, \emph{model soups} are obtained by linear interpolation (or extrapolation) of the adversarial and clean tokens. We notice that the clean and adversarial tokens are almost orthogonal ($\cos(\vphi_\textrm{clean},\vphi_\textrm{adv})=0.14$), so we can extend our study beyond \emph{model soups} by taking linear combinations of the two tokens $\beta_1 \vphi_\textrm{clean} + \beta_2\vphi_\textrm{adv}$. By taking a sweep over the $\beta_1$ and $\beta_2$ coefficients, we obtain in Figure \ref{fig:2d_landscape} the clean and robust accuracy landscapes in the plane defined by the two tokens and where the diagonal corresponds to the \emph{model soups}. We observe that the main direction of change for the clean and robust accuracies is the \emph{model soups} diagonal (top left to bottom right). We can clearly see the trade-off in clean/robust accuracy, but also there seems to be a compromise region between clean and robust accuracy as the other diagonal (from bottom left to top right) is visually distinct for clean and robust accuracy. In panel (c) of Figure \ref{fig:2d_landscape}, we plot the arithmetic mean between the normalized (with min/max rescaling) clean and robust accuracies. We observe that the best compromises between clean and robust accuracy have a stronger adversarial token weight than the clean token weight.

\section{Limitations}
We have empirically shown that co-training a fully shared \textsc{ViT} does not retain any robustness whereas having two classification tokens specific to the clean and adversarial images is enough to get competitive performance both in clean and robust accuracy. However, we leave to future work the theoretical explanation on why this small architecture change (adding only 768 parameters) results in such a gap in performance.
Similarly, beyond our intuition that parameter sharing when using adapters makes \emph{model soups} possible, we cannot support our empirical results with theory and leave it to future work. 

\section{Additional tables and figure}

In the following we present additional tables and figures of results described in the main part but omitted above because of space limits.

\begin{table}[h]
 \caption{{\bf Model soups for several weightings $\beta$.} We report the accuracy for various \emph{model soups} on \imagenet variants by interpolating between the clean and adversarial tokens with several weightings $\beta$. $\beta=1$ boils down to the \emph{clean mode} and $\beta=0$ to the \emph{robust mode}. \label{tab:soup_betas}}%
\begin{center}
\resizebox{1\textwidth}{!}{
\begin{tabular}{l|cccccccc|c}
    \hline
    \cellcolor{header} $\beta$ & \cellcolor{header} \textsc{ImageNet}  & \cellcolor{header} \textsc{IN-Real} & \cellcolor{header} \textsc{IN-V2} & \cellcolor{header} \textsc{IN-A} & \cellcolor{header} \textsc{IN-R} & \cellcolor{header} \textsc{IN-Sketch} & \cellcolor{header} \textsc{Conflict} & \cellcolor{header} \textsc{IN-C} & \cellcolor{header} \textsc{Mean}  \TBstrut \\
    \hline
    0 & 78.25\% & 84.73\% & 66.04\% & 13.39\% & 55.23\% & 39.62\% & \textbf{56.48\%} & 56.67\% & 56.30\%    \Tstrut \\
    0.1 & 78.63\% & 85.03\% & 66.49\% & 14.00\% & 55.31\% & 39.79\% & 55.70\% & 57.26\% & 56.53\%   \\
    0.2 & 79.14\% & 85.40\% & 67.13\% & 15.11\% & 55.23\% & 40.00\% & 54.37\% & 58.23\% & 56.83\%   \\
    0.3 & 79.74\% & 85.89\% & 67.95\% & 16.96\% & 55.40\% & 40.26\% & 53.36\% & 59.73\% & 57.41\%   \\
    0.4 & 80.50\% & 86.47\% & 68.91\% & 19.60\% & 55.55\% & 40.47\% & 51.09\% & 61.91\% & 58.06\%   \\
    0.5 & 81.49\% & 87.18\% & 70.21\% & 23.56\% & \textbf{55.64\%} & 40.70\% & 49.14\% & 64.74\% & 59.08\%   \\
    0.6 & 82.38\% & 87.76\% & 71.43\% & 28.63\% & 55.58\% & 40.98\% & 46.88\% & 67.52\% & 60.15\%   \\
    0.7 & 83.05\% & 88.26\% & 72.41\% & 33.72\% & 55.43\% & 41.19\% & 44.38\% & 69.27\% & 60.96\%   \\
    0.8 & 83.54\% & 88.46\% & 72.94\% & 36.79\% & 55.07\% & \textbf{41.22\%} & 41.64\% & 69.97\% & 61.20\%   \\
    0.9 & 83.69\% & 88.50\% & 73.48\% & 38.23\% & 54.74\% & 41.17\% & 40.00\% & \textbf{70.07\%} & \textbf{61.23\%}   \\
    1 & \textbf{83.76\%} & \textbf{88.52\%} & \textbf{73.53\%} & \textbf{38.37\%} & 54.43\% & 41.15\% & 39.77\% & 69.92\% & 61.18\%   \Bstrut  \\
    \hline
    \end{tabular}
}
\end{center}
\end{table}

\begin{table}[h]
 \caption{{\bf Ensembles for several weightings $\beta$.} We report the accuracy on \imagenet variants for various ensembles obtained by averaging the probability predictions of the \emph{clean and robust modes} of a \vitb with classification token adapter. Here $\beta$ is the weight used to average the probability predictions. $\beta=1$ boils down to the \emph{clean mode} and $\beta=0$ to the \emph{robust mode}. \label{tab:ensemble_betas}}%
\begin{center}
\resizebox{1\textwidth}{!}{
\begin{tabular}{l|cccccccc|c}
    \hline
    \cellcolor{header} $\beta$ & \cellcolor{header} \textsc{ImageNet}  & \cellcolor{header} \textsc{IN-Real} & \cellcolor{header} \textsc{IN-V2} & \cellcolor{header} \textsc{IN-A} & \cellcolor{header} \textsc{IN-R} & \cellcolor{header} \textsc{IN-Sketch} & \cellcolor{header} \textsc{Conflict} & \cellcolor{header} \textsc{IN-C} & \cellcolor{header} \textsc{Mean}  \TBstrut \\
    \hline
    0 & 78.25\% & 84.73\% & 66.04\% & 13.39\% & 55.23\% & 39.62\% & \textbf{56.48\%} & 56.67\% & 56.30\% \Tstrut  \\
    0.1 & 79.97\% & 86.10\% & 68.33\% & 21.23\% & 56.26\% & 40.15\% & 56.41\% & 64.14\% & 59.07\% \\
    0.2 & 81.26\% & 87.11\% & 70.03\% & 25.13\% & 56.56\% & 40.63\% & 55.08\% & 67.09\% & 60.36\%   \\
    0.3 & 82.15\% & 87.74\% & 71.24\% & 27.72\% & 56.63\% & 40.95\% & 53.67\% & 68.57\% & 61.08\%   \\
    0.4 & 82.81\% & 88.16\% & 72.09\% & 29.96\% & \textbf{56.68\%} & 41.11\% & 52.50\% & 69.36\% & 61.58\%   \\
    0.5 & 83.24\% & 88.39\% & 72.88\% & 32.13\% & 56.65\% & 41.28\% & 51.41\% & 69.79\% & 61.97\%   \\
    0.6 & 83.47\% & 88.51\% & 73.28\% & 33.64\% & 56.46\% & 41.35\% & 50.23\% & 69.98\% & 62.12\%   \\
    0.7 & 83.62\% & 88.58\% & 73.36\% & 35.05\% & 56.32\% & \textbf{41.36\%} & 49.14\% & \textbf{70.04\%} & \textbf{62.18\%}   \\
    0.8 & 83.68\% & 88.58\% & 73.37\% & 36.27\% & 56.03\% & 41.35\% & 46.64\% & \textbf{70.04\%} & 61.99\%   \\
    0.9 & 83.73\% & 88.57\% & 73.47\% & 37.25\% & 55.62\% & 41.28\% & 44.61\% & 69.99\% & 61.81\%   \\
  1 & \textbf{83.76\%} & \textbf{88.52\%} & \textbf{73.53\%} & \textbf{38.37\%} & 54.43\% & 41.15\% & 39.77\% & 69.92\% & 61.18\%   \Bstrut  \\
    \hline
    \end{tabular}
}
\end{center}
\end{table}

\begin{table}[h]
 \caption{We report the clean and robust accuracy for \emph{model soups} with various weightings $\beta$ of two independently trained \vitb, one nominally and the other adversarially. $\beta=0$ boils down to the model trained adversarially and $\beta=1$ to the model trained nominally. We notice that both robust and clean accuracies drop immediately when the weighting factor $\beta$ between parameters is not equal to 0 or 1. \label{tab:indep_soup}}%
\begin{center}
\resizebox{0.3\textwidth}{!}{
\begin{tabular}{l|cc}
    \hline
    \cellcolor{header} $\beta$ & \cellcolor{header} \textsc{Clean} & \cellcolor{header} \textsc{Robust}  \TBstrut \\
    \hline
    0 & 76.88\% & 56.19\%  \Tstrut \\
0.1 & 28.72\% & 8.56\%   \\
0.2 & 1.38\% & 0.06\%   \\
0.3 & 0.21\% & 0.00\%   \\
0.4 & 0.16\% & 0.00\%   \\
0.5 & 0.14\% & 0.01\%   \\
0.6 & 0.10\% & 0.01\%   \\
0.7 & 1.72\% & 0.09\%   \\
0.8 & 43.25\% & 0.14\%   \\
0.9 & 78.56\% & 0.01\%   \\
1 & 82.64\% & 0.00\%  \Bstrut \\
    \hline
    \end{tabular}
}
\end{center}
\end{table}

\begin{figure}
  \centering
  \includegraphics[width=0.5\linewidth]{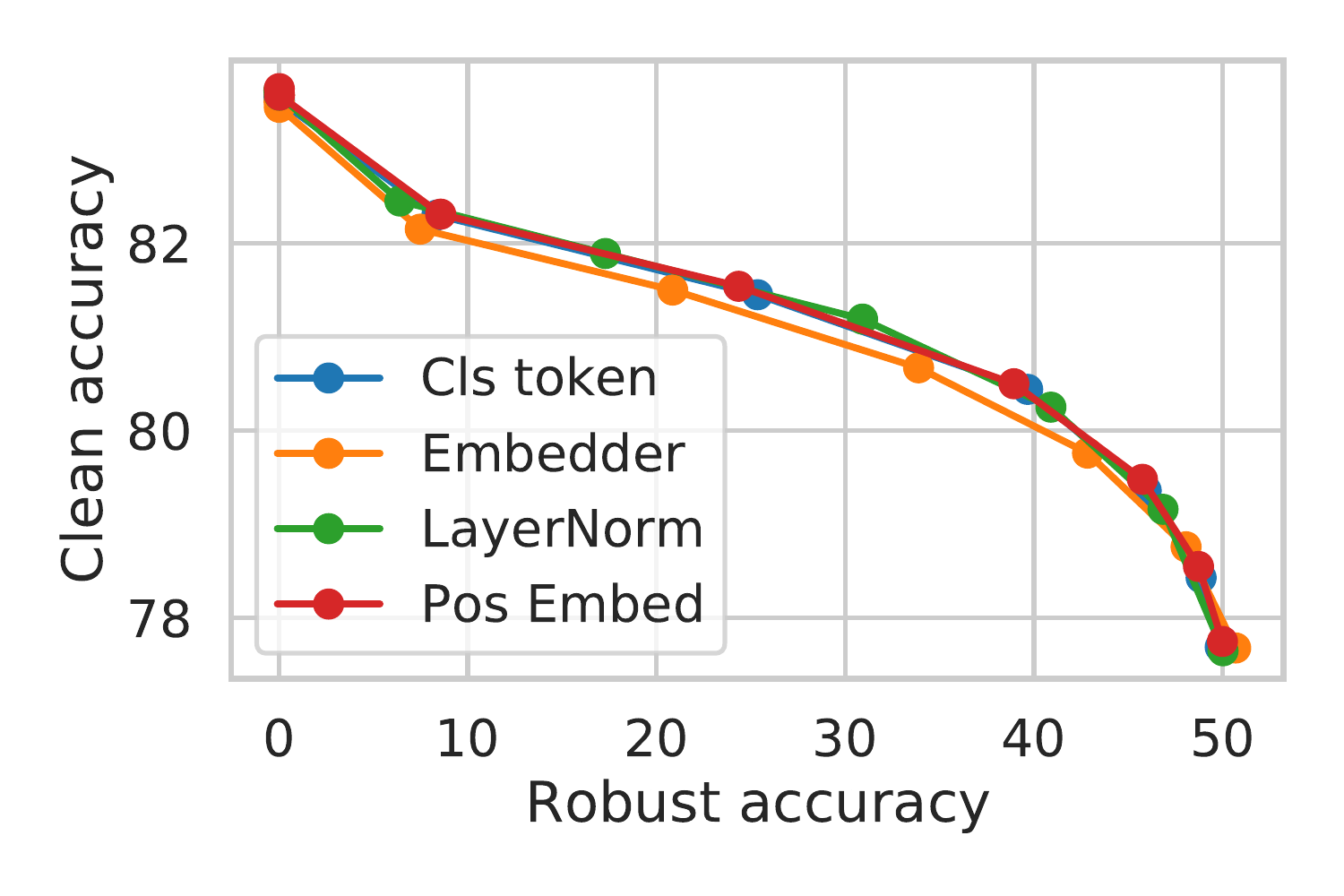}
  \caption{{\bf Comparing soups for various adapters.} We report the clean and robust accuracy (single mode for both) on \imagenet for \emph{model soups} of networks trained with various types of adapters. \label{fig:soup_adapters}}%
\end{figure}

\end{document}